\documentclass{article}

\PassOptionsToPackage{numbers, compress}{natbib}
\usepackage[preprint]{neurips_2025}

\usepackage[utf8]{inputenc} 
\usepackage[T1]{fontenc}    
\usepackage{hyperref}       
\usepackage{url}            
\usepackage{booktabs}       
\usepackage{amsfonts}       
\usepackage{nicefrac}       
\usepackage{microtype}      
\usepackage{xcolor}         
\usepackage{wrapfig}
\usepackage{comment}

\usepackage{microtype}
\usepackage{graphicx}
\usepackage{booktabs} 
\usepackage{subcaption}

\usepackage{hyperref}
\usepackage{balance}

\usepackage{amsmath}
\usepackage{amssymb}
\usepackage{mathtools}
\usepackage{amsthm}
\usepackage{algorithm}
\usepackage{algorithmic}

\usepackage[capitalize,noabbrev]{cleveref}

\theoremstyle{plain}
\newtheorem{theorem}{Theorem}[section]
\newtheorem{proposition}[theorem]{Proposition}

\theoremstyle{definition}

\theoremstyle{remark}

\usepackage[textsize=tiny]{todonotes}

\title{Multi-Objective Hyperparameter Selection via Hypothesis Testing on Reliability Graphs}

\author{%
  Amirmohammad Farzaneh \quad
  Osvaldo Simeone\\
Centre for Intelligent Information Processing Systems\\
Department of Engineering\\
King's College London\\
London, United Kingdom\\
  \texttt{\{amirmohammad.farzaneh,osvaldo.simeone\}@kcl.ac.uk} \\
}

\begin{document}

\maketitle

\begin{abstract}
The selection of hyperparameters, such as prompt templates in large language models (LLMs), must often strike a balance between reliability and cost. In many cases, structural relationships between the expected reliability levels of the hyperparameters can be inferred from prior information and held-out data -- e.g., longer prompt templates may be more detailed and thus more reliable. However, existing hyperparameter selection methods either do not provide formal reliability guarantees or are unable to incorporate structured knowledge in the hyperparameter space. This paper introduces reliability graph-based Pareto testing (RG-PT), a novel multi-objective hyperparameter selection framework that maintains formal reliability guarantees in terms of false discovery rate (FDR), while accounting for known relationships among hyperparameters via a directed acyclic graph. Edges in the graph reflect expected reliability and cost trade-offs among hyperparameters, which are inferred via the Bradley-Terry (BT) ranking model from prior information and held-out data. Experimental evaluations demonstrate that RG-PT significantly outperforms existing methods such as learn-then-test (LTT) and Pareto testing (PT) through a more efficient exploration of the hyperparameter space.

\end{abstract}

\section{Introduction}
\label{sec:intro}

\subsection{Context and Motivation}
\label{sec:context}

\begin{figure}
  \centering
  \begin{subfigure}{0.71\linewidth}
    \includegraphics[width=\linewidth]{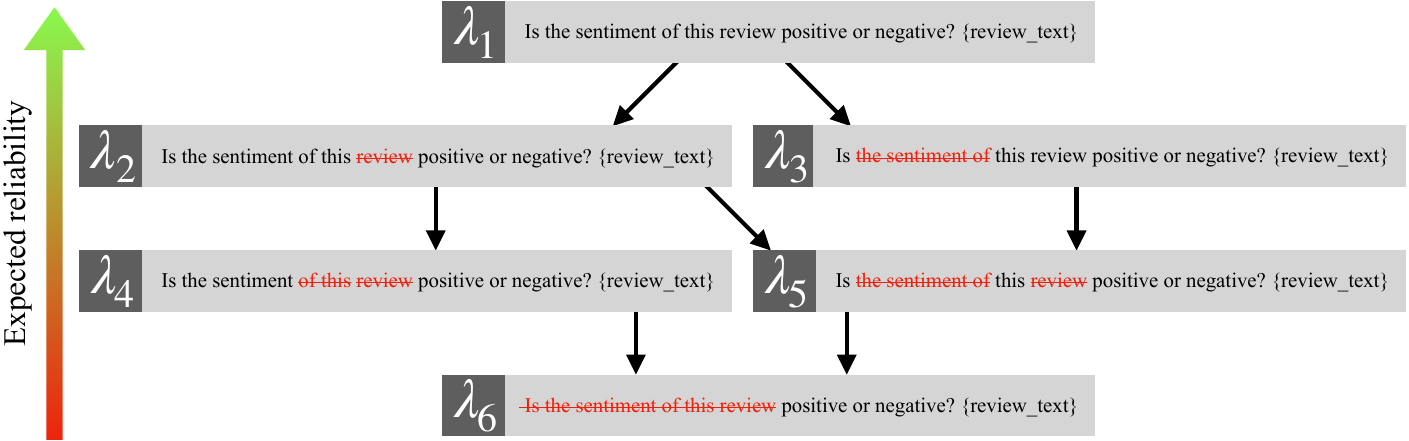}
    \caption{}
      \label{fig:intro_a}
  \end{subfigure}
  \hfill
  \begin{subfigure}{0.28\linewidth}
    \includegraphics[width=\linewidth]{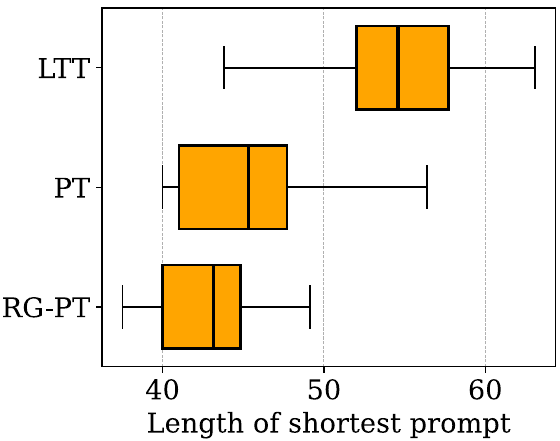}
    \caption{}
    \label{fig:intro_b}
  \end{subfigure}
\caption{Illustrative example for prompt engineering in LLM-based sentiment analysis: (a) 
 Prompt template candidates in set $\Lambda$ have expected reliability levels that can be arranged on a reliability graph (RG), so that  each parent prompt template is expected to be more reliable than its child prompts; (b)  Distribution of the length of the shortest prompt templates identified by LTT \cite{angelopoulos2021learn}, PT \cite{laufer2022efficiently}, and the proposed RG-PT for the Stanford Sentiment Treebank dataset \cite{socher2013recursive} (see Sec. \ref{sec:prompt_engineering} for details).}

  \label{fig:intro}
\end{figure}

Consider the problem of selecting prompt templates for a large language model (LLM)-based sentiment analysis task \cite{honovich2022instruction}. In this setting, the LLM receives a natural language prompt template along with a movie review as input, and the goal is to determine whether the sentiment expressed in the review is positive or negative. As illustrated in Fig. \ref{fig:intro}, the prompt templates $\lambda$ are chosen from a set of pre-determined choices  $\Lambda$, and the objective is to identify prompt templates $\lambda$ that elicit consistently accurate responses across inputs \cite{honovich2022instruction}. 

Longer prompts often yield more reliable outputs \cite{fu2023gptscore}. However,  they are also more costly to the end user when pay-per-token billing schemes are applied. This is commonly the case for enterprise software incorporating AI-driven analytics or hosted LLM endpoints via the LLM application programming interface (API)  \cite{chen2024frugalgpt}. As exemplified in Fig. \ref{fig:intro_a}, this suggests that the reliability of different prompt templates follows a \textit{directed acyclic graph} (DAG) structure with nodes closer to the roots corresponding to more costly prompt templates with a higher expected reliability.

An ideal prompt engineering scheme would apply \textit{hyperparameter selection} methods capable of selecting prompt templates that are as short as possible while enduring formal reliability guarantees. Existing hyperparameter selection methods, however, either do not meet formal reliability requirements \cite{bergstra2011algorithms, li2018hyperband}, or cannot incorporate the \emph{structured  knowledge} encoded in a graph like the DAG in Fig. \ref{fig:intro_a}.

In particular, \emph{Learn-Then-Test} (LTT) \cite{angelopoulos2021learn} pioneered the use of \textit{multiple hypothesis testing} (MHT) for hyperparameter selection, providing formal guarantees on the reliability of the returned subset of hyperparameters. However, LTT cannot incorporate structured information about relative expected  reliability levels of the hyperparameters. \emph{Pareto Testing} (PT) \cite{laufer2022efficiently} builds on LTT to address multi-objective optimization problems. Specifically, PT infers a global, \emph{linear} ordering over hyperparameters from held-out data based on their expected relative reliability. Thus, PT cannot account for more complex structured relationships between candidate hyperparameters as in the example of Fig. \ref{fig:intro_a}.

This paper proposes a novel framework, \textit{reliability graph-based PT} (RG-PT), that systematically captures and exploits interdependencies between hyperparameter configurations for hyperparameter selection. RG-PT models the hyperparameter space as a DAG, termed the \textit{reliability graph} (RG). In an RG,  nodes correspond to candidate hyperparameter configurations, such as prompt templates, and edges encode reliability relationships. If there is an edge from a hyperparameter $\lambda_i$ to another $\lambda_j$ in the RG, then $\lambda_i$ is expected to be more reliable than $\lambda_j$. 

For the running example of prompt design, as shown in Fig. \ref{fig:intro_b} (detailed in Sec. \ref{sec:experiments}), RG-PT is seen experimentally to select shorter prompt templates than LTT and PT, while still satisfying formal guarantees in terms of \emph{false discovery rate} (FDR). The FDR measures the fraction of unreliable prompt templates returned by the hyperparameter selection scheme. This advantage of RG-PT stems from its ability to encode rich structural relationships among prompt templates, so as to explore the hyperparameter space more efficiently during the MHT procedure.



\subsection{Further Related Work}

\textit{Hyperparameter Selection:} State-of-the-art techniques for hyperparameter optimization, such as Bayesian optimization \cite{snoek2012practical}, bandit-based methods \cite{li2018hyperband}, and gradient-based optimization \cite{maclaurin2015gradient}, provide satisfactory empirical performance, but they lack statistical guarantees. LTT addresses this gap by incorporating MHT in the hyperparameter selection process \cite{angelopoulos2021learn}.   Extensions of LTT are surveyed in \cite{farzaneh2025ensuring}. Note that, while reference \cite{angelopoulos2021learn} mentions the possible use of graph-based approaches, these are limited to fixed user-defined graphs or linear directed graphs (chains).


\textit{Multi-Objective Optimization:} Modern AI applications often require optimizing multiple objectives such as accuracy, efficiency, and cost. This can be formally done through Pareto optimization to identify all the feasible trade-off points among different objectives \cite{deb2002fast}. PT \cite{laufer2022efficiently} extends LTT to settings with multiple objectives, inferring a linear testing order in the hyperparameter space based on held-out data.


\subsection{Main Contributions}

The main contributions of this paper are as follows.

    \textbf{Methodology:} We propose RG-PT, a novel multi-objective hyperparameter selection framework that systematically infers and utilizes interdependencies among the expected reliability levels of candidate hyperparameter configurations. RG-PT first constructs a DAG, known as RG, based on prior information and held-out data via the Bradley-Terry (BT) ranking model \cite{hunter2004mm} and the non-negative Lasso \cite{tibshirani1996regression}. Then, it applies MHT-based hyperparameter selection on the RG by following DAGGER, a graphical testing method introduced in \cite{ramdas2017dagger}.
    
    \textbf{Applications:} We demonstrate the effectiveness of RG-PT through experiments in LLM prompt engineering, sequence-to-sequence translation, object detection, image classification, and telecommunications, highlighting its advantages over existing methods.

The rest of this paper is organized as follows. In Sec. \ref{sec:problem_formulation}, we define the multi-objective hyperparameter selection problem. Sec. \ref{sec:RG-PT} details the proposed RG-PT framework. Experimental results are presented in Sec. \ref{sec:experiments}. We conclude the paper in Sec. \ref{sec:conclusion}.

\section{Multi-Objective Hyperparameter Selection}
\label{sec:problem_formulation}

In this section, we define the problem of multi-objective hyperparameter selection, and we show how this problem can be formulated via MHT by following references \cite{angelopoulos2021learn, laufer2022efficiently}.

\subsection{Problem Definition}
\label{sec:problem_definition}

Consider a predefined discrete and finite set $\Lambda$ of hyperparameters $\lambda$, which govern the performance of a machine learning model such as an LLM (see, e.g., \cite{bergstra2012random}). The discrete set $\Lambda$ is populated in a preliminary pre-selection step \cite{eggensperger2021hpobench, falkner2018bohb, ijcai2021p296} using methods including LLM judges \cite{gu2024survey} and continuous optimizers like Bayesian optimization~\cite{bergstra2011algorithms} and Hyperband~\cite{li2018hyperband}.

In a multi-objective setting with $L$ risk functions, when tested on a data point $Z$, a hyperparameter $\lambda$ attains risk values $r_l(Z, \lambda)$ for $l = 1, \ldots, L$. The risk functions $r_l(Z,\lambda)$ are negatively oriented, meaning that lower risk values correspond to better-performing hyperparameters. The risks are normalized within the range $0 \leq r_l(Z, \lambda) \leq 1$. Furthermore, for each performance criterion $l = 1, \ldots, L$, the \textit{average risk function} is defined as
\begin{equation}
\label{eq:average_risk}
R_l(\lambda) = \mathbb{E}_{Z}\left[r_l(Z, \lambda)\right],
\end{equation}
where the expectation is taken over the distribution $P_Z$ of the data $Z$.

We partition the set of $L$ risk functions into the following two groups:

    1. \textit{Reliability risk functions: }The first set of risk functions $\{R_l(\lambda)\}_{l=1}^{L_c}$ must be controlled via the choice of the hyperparameter $\lambda$. In particular, a hyperparameter configuration $\lambda$ is said to be \textit{reliable} if it guarantees the constraints
    \begin{equation}
\label{eq:lambda_reliable}
R_l(\lambda) \leq \alpha_l \quad \text{for all}\quad l = 1, \ldots, L_c.
\end{equation}
    
    2. \textit{Auxiliary risk functions: }The second set of performance measures $\{R_l(\lambda)\}_{l=L_c+1}^{L}$ are unconstrained, and are optimized in a best-effort fashion via the selection of the hyperparameter $\lambda$.

Accordingly, the goal of hyperparameter selection is defined as the \textit{multi-objective problem}
\begin{equation}
\label{eq:goal}
\begin{aligned}
    &\min_{\lambda \in \Lambda} \; \{ R_{L_c+1}(\lambda), R_{L_c+2}(\lambda), \dots, R_L(\lambda) \} \\
    &\text{subject to} \; R_l(\lambda) < \alpha_l \; \text{for all} \; 1 \leq l \leq L_c,
\end{aligned}
\end{equation}

which targets the minimization of the auxiliary risk functions $\{R_l(\lambda)\}_{l = L_c + 1}^L$ under constraints on the reliability risk functions $\{R_l(\lambda)\}_{l = 1}^{L_c}$. For example, in the setting of Fig. \ref{fig:intro}, we wish to minimize the prompt length, while ensuring a constraint on the accuracy of the LLM's outputs.

Solving a multi-objective optimization problem such as (\ref{eq:goal}) ideally entails identifying the entire Pareto front of dominant solutions $\lambda \in \Lambda$, or at least obtaining specific solutions corresponding to scalar criteria \cite{deb2016multi, jamieson2016non}. However, the problem (\ref{eq:goal}) cannot be directly addressed since the data distribution $P_Z$ is assumed to be unknown. Instead, we assume to have access to i.i.d. data $\mathcal{Z} = \{Z_j\}_{j = 1}^{n}$ drawn from the unknown data distribution $P_Z$. For any data subset $\widetilde{\mathcal{Z}}\subseteq \mathcal{Z}$, the empirical estimate of risk function $R_l(\lambda)$ can be obtained as
\begin{equation}
\label{eq:empirical_estimate}
\hat{R}_l(\lambda|\widetilde{\mathcal{Z}}) = \frac{1}{|\widetilde{\mathcal{Z}}|} \sum_{Z\in \widetilde{\mathcal{Z}}} r_l(Z, \lambda).
\end{equation}

\subsection{Hyperparameter Selection as Multiple Hypothesis Testing}
\label{sec:MHT}

As proposed in \cite{angelopoulos2021learn}, hyperparameter selection can be formally addressed as an MHT problem. Accordingly, for each hyperparameter $\lambda \in \Lambda$, we define the null hypothesis $\mathcal{H}_{\lambda}$ that hyperparameter $\lambda$ violates the reliability constraints (\ref{eq:lambda_reliable}), i.e.,
\begin{equation}
\label{eq:combined_hypothesis}
\mathcal{H}_{\lambda} :\;\text{there exists} \; l \in \{1, \dots, L_c\} \text{ such that } R_l(\lambda) > \alpha_l.
\end{equation}
Thus, rejecting the null hypothesis \( \mathcal{H}_{\lambda} \) implies that hyperparameter $\lambda$ meets all the constraints in (\ref{eq:lambda_reliable}). A rejection is also referred to as a \textit{discovery}. A discovery is \textit{false} if the selected hyperparameter $\lambda$ is actually unreliable, satisfying the null hypothesis $\mathcal{H}_\lambda$.

Given a dataset $\widetilde{\mathcal{Z}}\subseteq \mathcal{Z}$, evidence against the reliability of each candidate hyperparameter $\lambda\in \Lambda$ can be measured by the p-value \cite{rice2007mathematical}
\begin{equation}
    \label{eq:general_pval}
    p_{\lambda,l}(\widetilde{\mathcal{Z}}) = \exp(-2|\widetilde{\mathcal{Z}}|(\alpha_l - \hat{R}_l(\lambda|\widetilde{\mathcal{Z}}))_+^2)
\end{equation}
for each reliability risk function $l = 1, \ldots, L_c$, yielding the combined p-value
\begin{equation}
\label{eq:combined_pval}
p_\lambda(\widetilde{\mathcal{Z}}) = \max_{1 \leq l \leq L_c} p_{\lambda,l}(\widetilde{\mathcal{Z}}).
\end{equation}
The quantity $p_\lambda(\widetilde{\mathcal{Z}})$ can be shown to be a valid p-value for the null hypothesis $\mathcal{H}_\lambda$ in (\ref{eq:combined_hypothesis}). Therefore, thresholding the statistic (\ref{eq:combined_pval}) yields a reliability test that meets type-I error probability constraints \cite{laufer2022efficiently}.

Furthermore, by formulating hyperparameter selection as an MHT problem, we can leverage statistical tools that guarantee \textit{false discovery rate} (FDR) requirements \cite{casella2024statistical}. To elaborate, define as $\hat{\Lambda}_\mathcal{Z}$ the subset of hyperparameters selected by an MHT mechanism. The FDR is defined as the expected proportion of unreliable hyperparameters in set $\hat{\Lambda}_\mathcal{Z}$. Therefore, controlling the FDR amounts to finding a subset $\hat{\Lambda}_\mathcal{Z}\subseteq \Lambda$ that satisfies the inequality
\begin{equation}
\label{eq:FDR_statistical}
    \mathbb{E}_Z\left[\frac{\sum_{\lambda \in \hat{\Lambda}_\mathcal{Z}} \mathbf{1}\{ R_l(\lambda) > \alpha_l \text{ for any } l = 1, \ldots, L_c \}}{\max(|\hat{\Lambda}_\mathcal{Z}|, 1)}\right] \leq \delta,
\end{equation}
where $\mathbf{1}\{\cdot\}$ is the indicator function, and the expectation is taken over the unknown data distribution $P_Z$. The FDR constraint (\ref{eq:FDR_statistical}) ensures that the average fraction of unreliable hyperparameters in set $\hat{\Lambda}_\mathcal{Z}$ is upper bounded by $\delta$.

\section{Reliability Graph-Based Pareto Testing}
\label{sec:RG-PT}

 In this section, we introduce RG-PT, a novel hyperparameter selection strategy based on MHT that adopts a testing schedule based on the novel concept of RG.

\subsection{Overview}

The design of RG-PT starts from the observation that the reliability of some hyperparameters can be highly predictive of the reliability of other hyperparameters, and that this structure can be encoded by a DAG as in Fig. \ref{fig:intro_a}. By incorporating the DAG structure in the MHT process of hyperparameter selection, RG-PT supports a more efficient hyperparameter selection procedure, while meeting formal reliability constraints in terms of the FDR (\ref{eq:FDR_statistical}).

\begin{wrapfigure}{r}{0.5\linewidth}
    \centering
    \includegraphics[width=\linewidth]{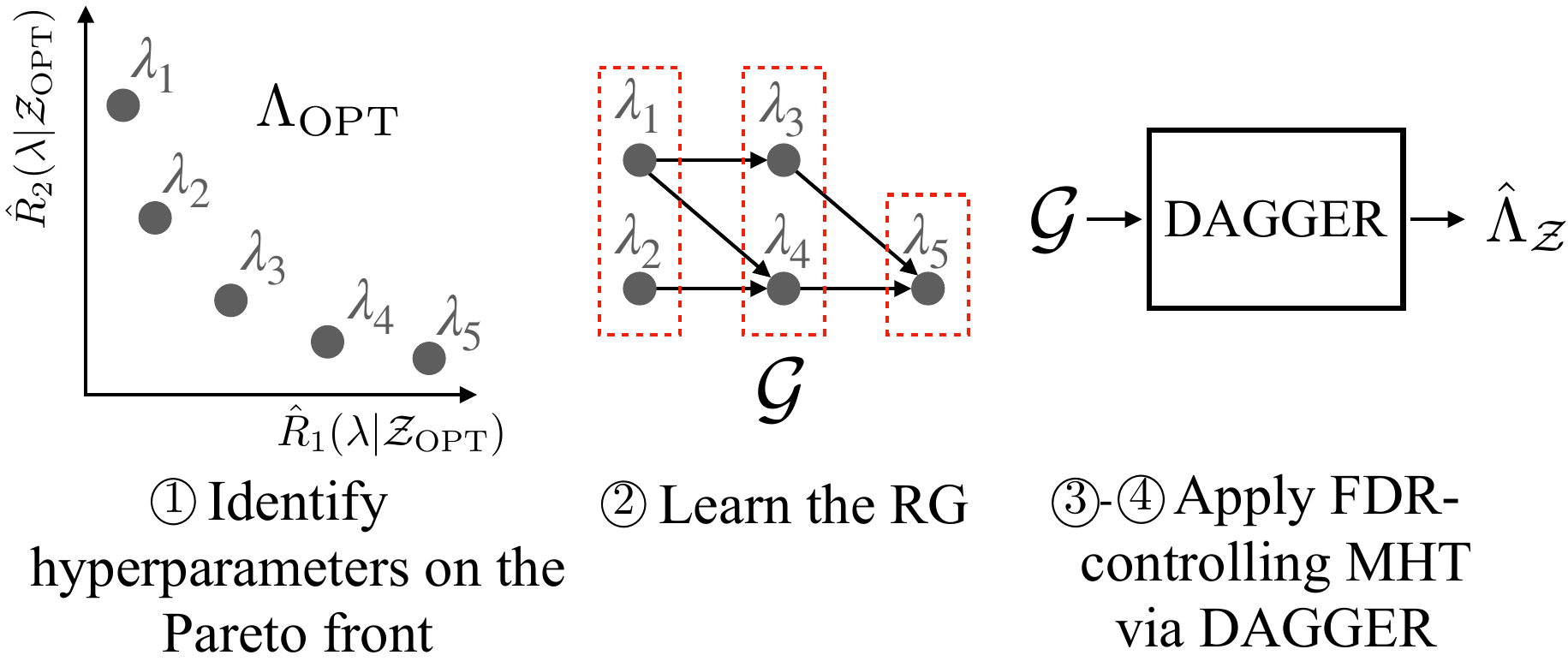}
    \caption{Illustration of the main steps of RG-PT: \textcircled{1} Estimate the hyperparameters $\Lambda_\text{OPT}$ lying on the Pareto front pf problem (\ref{sec:problem_formulation}); \textcircled{2} Build the RG over the selected hyperparameters $\Lambda_\text{OPT}$; \textcircled{3} Apply an FDR-controlling MHT procedure, DAGGER, to the RG to obtain the selected set $\hat{\Lambda}_\mathcal{Z}\subseteq \Lambda_\text{OPT}$.}
    \label{fig:DAG_PT}
\end{wrapfigure}

 As illustrated in Fig. \ref{fig:DAG_PT}, using a partition $\mathcal{Z} = \{\mathcal{Z}_\text{OPT}, \mathcal{Z}_\text{MHT}\}$ of the data set $\mathcal{Z}$, RG-PT applies the following steps:

     \textcircled{1} \textit{Estimating the Pareto front for all risk measures:} Following PT \cite{laufer2022efficiently}, RG-PT uses the dataset $\mathcal{Z}_\text{OPT}$ to identify the subset $\Lambda_\text{OPT}\subseteq \Lambda$ of hyperparameters that are on the Pareto front of the space of estimated risk measures $\{\hat{R}_l(\lambda|\mathcal{Z}_\text{OPT})\}_{l=1}^L$. This is done by addressing the multi-objective optimization problem (\ref{eq:goal}) with the estimates $\{\hat{R}_l(\lambda|\mathcal{Z}_\text{OPT})\}_{l=1}^L$ in lieu of the true risks $\{R_l(\lambda|\mathcal{Z}_\text{OPT})\}_{l=1}^L$ using any suitable multi-objective optimization algorithm \cite{laufer2022efficiently}.

    \textcircled{2} \textit{Learning the reliability graph:} Rather than ordering the hyperparameters in subset $\Lambda_\text{OPT}$ in a linear sequence as done by PT \cite{laufer2022efficiently}, RG-PT creates an RG, with nodes given by the hyperparameters in subset $\Lambda_\text{OPT}$. This is done by following the principle that hyperparameters $\lambda \in \Lambda_\text{OPT}$ whose reliability levels are predictive of the reliability levels of other hyperparameters $\Lambda' \subset \Lambda_\text{OPT}$ should be tested before the hyperparameters $\Lambda'$. As detailed in Sec. \ref{sec:DAG}, the RG construction leverages the BT model to incorporate prior information and the non-negative Lasso to determine the links in the graph.

    \textcircled{3} \textit{FDR-controlling MHT:} Using the data set $\mathcal{Z}_\text{MHT}$, FDR-controlling MHT is carried out by incorporating the structure encoded by the RG. As explained in Sec. \ref{sec:DAG}, this is done by using DAGGER \cite{ramdas2017dagger}, returning the subset $\hat{\Lambda}_\mathcal{Z} \subseteq \Lambda_\text{OPT}$.

    \textcircled{4} \textit{Addressing the multi-objective optimization problem}: Given the subset $\hat{\Lambda}_\mathcal{Z}$, RG-PT addresses the problem 
\begin{equation}
\label{eq:multi_mini}
    \min_{\lambda \in \hat{\Lambda}_{\mathcal{Z}}} \{\hat{R}_{L_c+1}(\lambda|\mathcal{Z}_\text{OPT}),\ldots, \hat{R}_{L}(\lambda|\mathcal{Z}_\text{OPT})\},
\end{equation}
where the auxiliary risk functions $R_l(\lambda)$ in (\ref{eq:goal}) are replaced with the corresponding empirical estimates (\ref{eq:empirical_estimate}) obtained with data set $\mathcal{Z}_\text{OPT}$.

\subsection{Learning the Reliability Graph}
\label{sec:DAG}

 After obtaining the estimated Pareto front $\Lambda_{\text{OPT}}$, RG-PT constructs an RG to encode the expected relationships between the reliability levels attained by the candidate hyperparameters in the set $\Lambda_\text{OPT}$. The RG is a DAG in which each node represents a hyperparameter $\lambda \in \Lambda_\text{OPT}$, and edges are directed to describe a reliability hierarchy. Specifically, edges encode the expectation that parent nodes are predictive of the reliability of their child nodes.

Accordingly, starting from the nodes with no parents and following the direction of the edges in the RG, one encounters hyperparameters that are estimated to be increasingly unreliable. Generalizing the linear ordering assumed by PT, the DAG structure adopted by RG-PT can thus assign the same expected reliability ranking to multiple hyperparameters. Specifically, all the hyperparameters at the same depth in the DAG are deemed to have the same relative reliability level. This partial ordering enables a more efficient exploration of the hyperparameter space.

In order to learn the RG, RG-PT leverages the data set $\mathcal{Z}_\text{OPT}$, as well as, possibly, prior information about the relative reliability of pairs of hyperparameters. This is done via the following two steps:

      1. \textit{Depth assignment:} The hyperparameters in set $\Lambda_\text{OPT}$ are ranked in terms of their expected reliability, allowing for multiple hyperparameters to be ranked equally. This step thus assigns a depth level $d$ in the DAG to each hyperparameter $\lambda \in \Lambda_\text{OPT}$. As explained in Sec. \ref{sec:depth_assignment}, this is done by leveraging the BT ranking model \cite{hunter2004mm}.
    
    2. \textit{Learning the directed edges:} Given any hyperparameter $\lambda$ at some depth $d$, RG-PT selects a subset of hyperparameters at the previous depth level $d-1$ to serve as parents of the hyperparameter $\lambda$. As specified in Sec. \ref{sec:learning_edges}, this is done by choosing the hyperparameters at depth $d-1$ that are most predictive of the reliability level of hyperparameter $\lambda$ via the non-negative Lasso \cite{tibshirani1996regression}.

\subsubsection{Depth Assignment}
\label{sec:depth_assignment}

 Fix a number $D\leq |\Lambda_\text{OPT}|$ of levels for the DAG. With $D = |\Lambda_\text{OPT}|$, one can assign each hyperparameter $\lambda \in \Lambda_\text{OPT}$ a distinct level, yielding a global ordering and recovering PT. Conversely, with $D=1$, all hyperparameters $\lambda \in \Lambda_\text{OPT}$ are assigned to the same level. The setting of interest is thus $1 < D < |\Lambda_\text{OPT}|$, which is assumed from now on. 

 Depth assignment is carried out by first obtaining a score $s(\lambda)$ for all hyperparameters $\lambda \in \Lambda_\text{OPT}$ using the data set $\mathcal{Z}_\text{OPT}$, and then partitioning the set $\Lambda_\text{OPT}$ into $D$ clusters according to the obtained scores.

To compute the scores \( s(\lambda) \) for hyperparameters \( \lambda \in \Lambda_\text{OPT} \), we use the BT model \cite{hunter2004mm}. The BT model converts pairwise counts $w_{ij}$ for all pairs of hyperparameters $\lambda_i$ and $\lambda_j$ in subset $\Lambda_{\text{OPT}}$ into per-hyperparameter scores $s(\lambda)$ for all $\lambda \in \Lambda_\text{OPT}$. The pairwise count $w_{ij}$ measures the number of times that hyperparameter $\lambda_i$ was found to be more reliable than hyperparameter $\lambda_j$. In RG-PT, we propose to evaluate the pairwise counts $w_{ij}$ by leveraging two sources of information:

$\bullet$ \textit{Prior information:} Prior information is encoded by pairwise probabilities $0\leq \eta_{ij}\leq 1$ for each pair of hyperparameters $(\lambda_i, \lambda_j) $ in $ \Lambda_\text{OPT}$. This probability reflects the expected rate at which hyperparameter $\lambda_i$ is observed to be more reliable than hyperparameter $\lambda_j$. Note that we have $\eta_{ji} = 1- \eta_{ij}$. The strength of the prior information is determined via a pseudocount variable $n_p$ as in the standard categorical-Dirichlet model \cite{bishop2006pattern}. A larger pseudocount $n_p$ indicates a stronger trust in the prior information. Importantly, the statistical guarantees of RG-PT do not depend on the choice of the prior probabilities $\{\eta_{ij}\}$ and pseudocount $n_p$, which can, however, improve the capacity of RG-PT to optimize the auxiliary risk functions. Note that in the absence of prior information, one can set $n_p = 0$.

$\bullet$ \textit{Data:} Using the p-values $p_{\lambda_i}(\mathcal{Z}_\text{OPT})$ and $p_{\lambda_j}(\mathcal{Z}_\text{OPT})$ in (\ref{eq:combined_pval}), we evaluate the data-driven probability
\begin{equation}
\label{eq:data-driven_probs}
    p_{ij}(\mathcal{Z}_\text{OPT}) = \frac{p_{\lambda_i}(\mathcal{Z}_\text{OPT})}{p_{\lambda_i}(\mathcal{Z}_\text{OPT}) + p_{\lambda_j}(\mathcal{Z}_\text{OPT})}
\end{equation}
that hyperparameter $\lambda_i$ is more reliable than hyperparameter $\lambda_j$. Note that we have $p_{ij}(\mathcal{Z}_\text{OPT}) = 1- p_{ji}(\mathcal{Z}_\text{OPT})$.

Overall, the pair-wise count $w_{ij}$ is obtained by combining prior information and data as 
\begin{equation}
\label{eq:wij}
w_{ij} = |\mathcal{Z}_\text{OPT}| p_{ij}(\mathcal{Z}_\text{OPT}) + n_p \eta_{ij},
\end{equation}
so that the relative strength of the prior information in (\ref{eq:wij}) depends on the ratio $n_p/|\mathcal{Z}_\text{OPT}|$ between the pseudocount $n_p$ and the number of data points $|\mathcal{Z}_\text{OPT}|$.

Using the BT model, the scores \( s(\lambda_i) \) for all hyperparameters $\lambda_i \in \Lambda_\text{OPT}$ are obtained by maximizing the log-likelihood \cite{hunter2004mm}  
\begin{equation}
\label{eq:BT_log}
\sum_{i=1}^{|\Lambda_\text{OPT}|} \sum_{j=1}^{|\Lambda_\text{OPT}|} \left( w_{ij} \ln \left(\frac{s(\lambda_i)}{s(\lambda_i) + s(\lambda_j)}\right) \right),
\end{equation}
with \( w_{ii} = 0 \) for all \( 1 \leq i \leq |\Lambda_\text{OPT}| \). With this design, in the absence of prior information (\( n_p = 0 \)), the BT model reduces to assigning scores directly proportional to the \( p \)-values \( p_\lambda(\mathcal{Z}_\text{OPT}) \).

 After obtaining the scores $s(\lambda_i)$ for all $1\leq i \leq |\Lambda_\text{OPT}|$, depth assignment is done via clustering, producing disjoint subsets $\Lambda_1, \ldots, \Lambda_D$. The cluster $\Lambda_1 \subseteq \Lambda_\text{OPT}$ contains the hyperparameters with the highest expected reliability, and the remaining clusters $\Lambda_2, \ldots, \Lambda_D$ are sorted in descending order of expected reliability. All hyperparameters in cluster $\Lambda_d$ are assigned depth level $d$.

 Clustering can be implemented by using methods such as $K$-means or hierarchical clustering. We recommend using agglomerative hierarchical clustering, which begins with each hyperparameter in its own cluster and iteratively merges clusters \cite{jain1988algorithms}.

\subsubsection{Learning the Directed Edges}
\label{sec:learning_edges}

 Having obtained the clusters $\Lambda_1, \ldots, \Lambda_D$, the RG is constructed by: (\textit{i}) including one node for each hyperparameter $\lambda \in \Lambda_\text{OPT}$; and (\textit{ii}) selecting for each hyperparameter $\lambda \in \Lambda_d$ at depth level $d$ a subset of hyperparameters in cluster $\Lambda_{d-1}$ to serve as parents of $\lambda$ for all depth levels $2\leq d\leq K$. The resulting directed edges are intended to represent inferred reliability dependencies.

 To this end, we implement feature selection via the non-negative Lasso \cite{tibshirani1996regression}. Specifically, given hyperparameter $\lambda \in \Lambda_d$, we consider the problem of predicting the risks $\{r_l(Z, \lambda)\}_{l = 1}^L$ from the risks $\{r_l(Z, \lambda')\}_{l =1}^{L}$ attained by the hyperparameters $\lambda \in \Lambda_{d-1}$ at the previous depth level. The use of non-negative Lasso regression ensures that only positive correlations are represented in the DAG, preserving hierarchical reliability relationships between parent and child nodes.

 Formally, using the data set $\mathcal{Z}_\text{OPT}$, for each hyperparameter $\lambda \in \Lambda_d$ we address the problem
\begin{equation}
\label{eq:Lasso}
\min_{\beta \geq 0}  \sum_{Z\in \mathcal{Z}_\text{OPT}} 
\left\|r(Z, \lambda) - \sum_{\lambda' \in \Lambda_{d-1}} \beta_{\lambda'} r(Z, \lambda')\right\|_2^2  + \tau \sum_{\lambda'\in \Lambda_{d-1}} \beta_{\lambda'},
\end{equation} where $\beta = \{\beta_{\lambda'}\}_{\lambda ' \in \Lambda_{d-1}}$ is the vector of non-negative regression coefficients corresponding to each potential parent node $\lambda' \in \Lambda_{d-1}$; $r(Z, \lambda)$ is the vector containing the values $\{r_l(Z,\lambda)\}_{l=1}^{L_c}$; $\|\cdot\|_2$ represents the $\ell_2$ norm; and $\tau > 0$ is a regularization parameter that controls the degree of sparsity in the solution. After solving the convex problem (\ref{eq:Lasso}), only the hyperparameters $\lambda' \in \Lambda_{j-1}$ for which the corresponding coefficient $\beta_{\lambda'}$ are positive are selected as parent nodes of hyperparameter $\lambda$. As for the variables $(n_p, \{\eta_{ij}\})$ in the BT likelihood (\ref{eq:BT_log}), the choice of the parameter $\tau$ does not affect the validity properties of RG-PT.

\subsection{FDR-Controlling Multiple Hypothesis Testing}
\label{sec:DAGGER2}

 Given the obtained RG, RG-PT performs MHT via DAGGER \cite{ramdas2017dagger}, an FDR-controlling algorithm that operates on DAGs. DAGGER begins testing at the root nodes of the RG, i.e., at the hyperparameters in cluster $\Lambda_1$, and proceeds with the clusters $\Lambda_2, \Lambda_3, \ldots, \Lambda_D$, guided by the outcomes of prior tests. If a hyperparameter is deemed unreliable, none of its descendants are tested.

The test for each hyperparameter $\lambda_i \in \Lambda_\text{OPT}$ detects $\lambda_i$ as reliable if the p-value $p_{\lambda_i}(\mathcal{Z}_\text{MHT})$ in (\ref{eq:combined_pval}) is no larger than a threshold $\delta_i$, i.e., if $p_{\lambda_{i}}(\mathcal{Z}_\text{MHT}) \leq \delta_i$. The testing level \(\delta_i\) is determined by DAGGER based on several factors, including the overall target FDR level \(\delta\) in constraint (\ref{eq:FDR_statistical}), the number of reliable hyperparameters identified among those tested prior to \(\lambda_i\), and the structure of the graph rooted at $\lambda_i$. We refer the reader to Appendix \ref{appendix:DAGGER} and to \cite{ramdas2017dagger} for details.

An algorithmic overview of RG-PT is provided in Appendix \ref{appendix:algorithm}. The following proposition states the theoretical guarantees provided by RG-PT.
\begin{proposition}
\label{prop:prop}
The set $\hat{\Lambda}_\mathcal{Z}$ of hyperparameters returned by RG-PT controls the FDR below the pre-specified threshold $\delta$ as in (\ref{eq:FDR_statistical}).
\end{proposition}

\begin{proof}
RG-PT applies the DAGGER algorithm \cite{ramdas2017dagger} to a DAG over the candidate hyperparameters, using valid p-values defined in (\ref{eq:combined_pval}). Since DAGGER controls the FDR at level $\delta$ for \textit{any} DAG structure when supplied with valid p-values under arbitrary dependence (see Appendix \ref{appendix:DAGGER}), the result follows directly.
\end{proof}

\section{Experiments}
\label{sec:experiments}

In this section, we evaluate the proposed RG-PT hyperparameter selection strategy on a prompt engineering problem \cite{honovich2022instruction} and a sequence-to-sequence translation task \cite{peters2021smoothing}. Additional experiments including on object detection \cite{angelopoulos2021learn} and telecommunications \cite{Nokia} can be found in Appendix \ref{appendix:experiments}\footnote{The code for the experiments can be found at the Github repository \href{https://github.com/kclip/RG-PT}{https://github.com/kclip/RG-PT}}.

Throughout the experiments, we adopt as benchmarks LTT \cite{angelopoulos2021learn} and PT \cite{laufer2022efficiently}. To the best of our knowledge, LTT and PT are the only existing hyperparameter selection methods that guarantee statistical validity in the sense of the FDR constraint (\ref{eq:FDR_statistical}), justifying this choice. LTT is implemented by applying Benjamini-Hochberg (BH) \cite{benjamini1995controlling} as the FDR-controlling algorithm, while PT follows \cite{laufer2022efficiently}, with the caveat that FDR-controlling FST \cite{lynch2017control} is used in lieu of an FWER-controlling scheme. LTT uses the entire calibration data set $\mathcal{Z}$ to evaluate the p-values used in BH, while PT and RG-PT partition $\mathcal{Z}$ into data sets $\mathcal{Z}_\text{OPT}$ and $\mathcal{Z}_\text{MHT}$.

\subsection{Reliable Prompt Engineering}
\label{sec:prompt_engineering}

\vspace{0.2em}
\noindent\textbf{Problem Setup.}
In this experiment, we focus on prompt engineering for the following three tasks from the instruction induction data set \cite{honovich2022instruction}: 1. \textit{Sentiment analysis:} In this task, based on the Stanford Sentiment Treebank dataset \cite{socher2013recursive}, each data point $Z = (X, Y)$ encompasses a movie review $X$, and the corresponding sentiment $Y\in\{\text{positive}, \text{negative}\}$. 2. \textit{Sentence similarity:} In this task, based on the Semantic Textual Similarity Benchmark dataset \cite{cer2017semeval}, each data point $Z = (X, Y)$ comprises two sentences as input $X$, along with a semantic similarity label $Y \in \{0,\dots,5\}$. 3. \textit{Word in context:} In this task, based on the Word-in-Context dataset \cite{pilehvar2018wic}, each data point $Z = (X, Y)$ consists of a target word and two context sentences as input $X$, paired with a binary label $Y \in \{\text{same}, \text{not same}\}$ indicating whether the target word shares the same meaning across both contexts.

For each task, we use 1000 examples each for the data sets $\mathcal{Z}_{\text{OPT}}$ and $\mathcal{Z}_{\text{MHT}}$, as well as for the test data set. Furthermore, following the forward generation mode detailed in \cite{zhou2022large}, we use the LLaMA3-70B-Instruct model \cite{llamma} to generate a set $\Lambda = \{\lambda_1, \dots, \lambda_{100}\}$ of distinct instruction-style prompt templates for each task.

Given a prompt $\lambda$ and an input $X$, the smaller LLaMA3-8B-Instruct model \cite{dubey2024llama} $f$ is queried with the concatenated input $[\lambda, X]$, producing the output $f([\lambda, X])$. For each input-output pair $Z = (X, Y)$, a task-specific 0-1 prompt loss is calculated as $r_\text{prompt}(Z, \lambda) = l(f([\lambda, X]), Y)\in \{0,1\}$, indicating whether the task was performed correctly. The objective is to find prompts in set $\Lambda$ that control the average prompt loss $R_\text{prompt}(\lambda) = \mathbb{E}_\mathcal{Z}\left[r_\text{prompt}(Z, \lambda)\right]$ below a target level of $\alpha = 0.2$, while minimizing the average prompt length. For this selection, we wish to control the FDR in (\ref{eq:FDR_statistical}) at level $\delta = 0.1$.


\noindent\textbf{Prior Information via LLM-as-a-Judge.} To incorporate prior structure into the reliability graph, we adopt the LLM-as-a-judge framework \cite{gu2024survey}. For each pair of prompts $\lambda_i, \lambda_j \in \Lambda_\text{OPT}$, we query the GPT-4 Turbo (gpt-4-0125-preview, $\text{temperature} = 0$, $\text{max tokens} = 10$) model \cite{openai2023gpt4turbo} with a task-specific prompt template to assess which instruction is more likely to elicit a correct or helpful response. We perform this comparison once per hyperparameter pair $\lambda_i, \lambda_j\in \Lambda_\text{OPT}$, and define the binary pairwise preference $\eta_{ij} = 1$ if GPT-4 Turbo selects $\eta_{ij}$ as more reliable, and $\eta_{ij} = 0$ otherwise. For each reliable set of prompts returned by each method, we choose the shortest prompt as the final hyperparameter choice. We set the pseudocount $n_p$ to 1,000.

\begin{figure}[h!]
  \centering
  \begin{subfigure}{0.4\linewidth}
    \includegraphics[width=\linewidth]{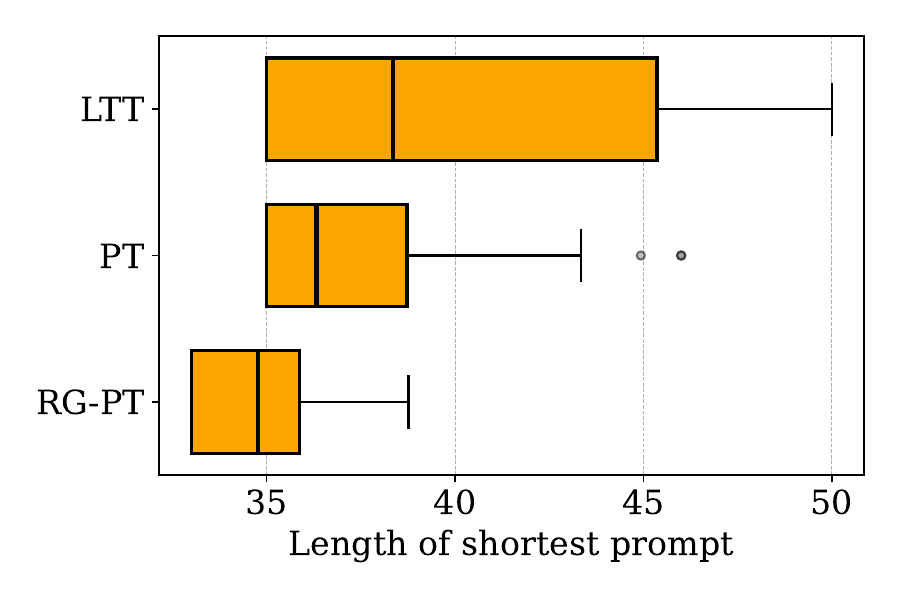}
    \caption{}
      \label{fig:task_a}
  \end{subfigure}
  \begin{subfigure}{0.4\linewidth}
    \includegraphics[width=\linewidth]{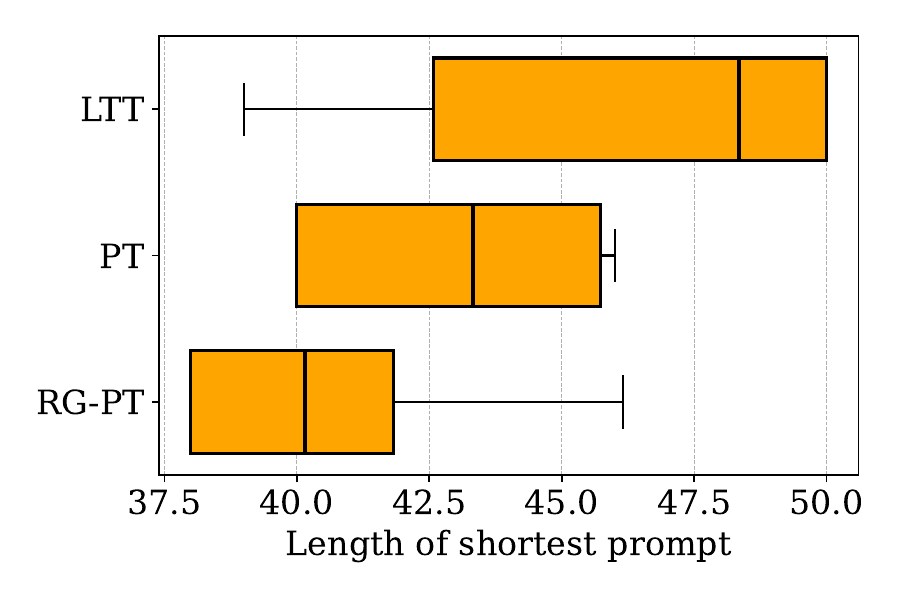}
    \caption{}
    \label{fig:task_b}
  \end{subfigure}
\caption{Distribution of the length of the shortest prompt templates identified by LTT \cite{angelopoulos2021learn}, PT \cite{laufer2022efficiently}, and the proposed RG-PT for (a) the Semantic Textual Similarity Benchmark dataset \cite{cer2017semeval} and (b) the Word-in-Context dataset \cite{pilehvar2018wic}.}
\label{fig:prompt}
\end{figure}

\noindent\textbf{Results.} For each task, we plot the distribution of the length of the shortest reliable prompt for 100 independent runs over random splits of the data set, for LTT, PT, and RG-PT in Fig.~\ref{fig:intro_b}, Fig. \ref{fig:task_a}, and Fig. \ref{fig:task_b} for the sentiment analysis task, the sentence similarity task, and the word in context task, respectively. The figures demonstrate that RG-PT identifies more concise instructions compared to LTT and PT by leveraging the prior information provided by the LLM judge. Note that all schemes satisfy the FDR constraint (not shown). For instance, RG-PT achieved an average FDR of 0.089, 0.095, and 0.092 for the the sentiment analysis, the sentence similarity, and the word in context tasks, respectively.

An ablation study on the effect of the RG depth $D$, as well as the effect of a misspecified prior information for this experiment can be found in Appendix \ref{appendix:ablation}.

\subsection{Sequence-to-Sequence Language Translation}
\label{sec:seq2seq_experiment}

We consider a sequence-to-sequence language translation task on the WMT16 Romanian-English dataset \cite{bojar-etal-2016-findings}, using BLEU \cite{papineni2002bleu} and ROUGE-L \cite{lin2004rouge} as the objectives. Following \cite{peters2021smoothing}, the dataset is preprocessed with SentencePiece tokenization \cite{kudo2018subword}, and an LSTM-based encoder-decoder is trained. Two key hyperparameters are considered:

1. The hyperparameter $\rho$ controls the \textit{sparsity} of the output distribution using Entmax \cite{peters2019sparse}, transitioning between a dense output with softmax ($\rho=1$) and sparsemax ($\rho=2$) \cite{martins2016softmax}.

2. The \textit{Fenchel-Young label smoothing strength} $\epsilon$ is a training regularization hyperparameter that determines the extent to which one-hot targets are mixed with uniform noise based on Fenchel-Young losses \cite{peters2021smoothing}. Accordingly, the original one-hot targets are assigned weight $1-\epsilon$, while the uniform distribution over all possible classes is assigned the weight $\epsilon$.

To create the initial candidate set $\Lambda$, we selected hyperparameters over a grid of 32 combinations, using 8 logarithmically spaced values in the interval $[1, 2]$ for $\rho$, and values in the set $\{0.0, 0.01, 0.05, 0.1\}$ for $\epsilon$. This selection is in line with reference \cite{peters2021smoothing}.

To set up RG-PT, we leveraged the prior knowledge that less sparse settings may be more reliable than their sparser counterparts. Specifically, for any two hyperparameters \(\lambda_i = (\rho_i, \epsilon_i)\) and \(\lambda_j = (\rho_j, \epsilon_j)\) where \(\rho_i < \rho_j\), we assigned a prior probability \(\eta_{ij} = 1\), reflecting this prior reliability assumption. Furthermore, the pseudocount parameter \(n_p\), which determines the weight of prior information, was set to be equal to \(|\mathcal{Z}_\text{OPT}|\).

\begin{wrapfigure}{r}{0.5\linewidth}
    \centering
    \includegraphics[width=\linewidth]{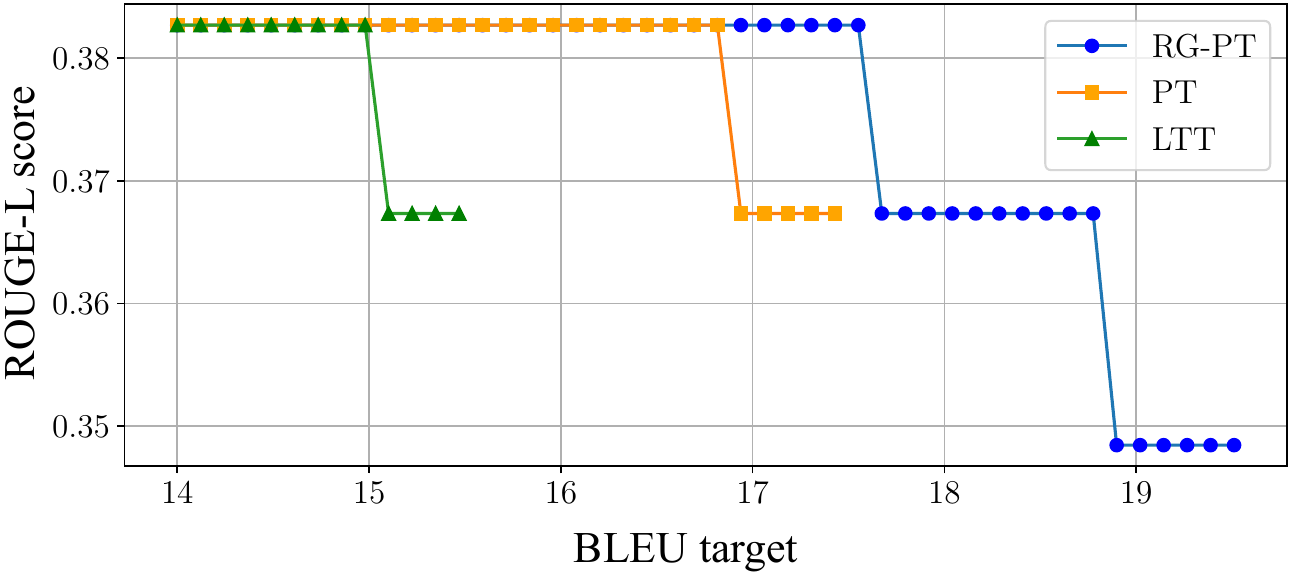}
    \caption{Test ROUGE-L scores achieved by LTT, PT, and RG-PT methods as a function of the target reliability value for the BLEU score.}
    \label{fig:seq2seq}
\end{wrapfigure}

Denote as $R_\text{BLEU}(\lambda)$ and $R_\text{ROUGE}(\lambda)$ the average BLEU and ROUGE-L scores, respectively, obtained for a given hyperparameter configuration $\lambda = (\rho, \epsilon)$. The goal is to guarantee the BLEU score to be above a threshold $\alpha$, while maximizing the ROUGE-L score. This amounts to an instance of problem (\ref{eq:goal}), with $L = 2$, $L_c = 1$, $R_1(\lambda) = -R_\text{BLEU}(\lambda)$, $R_2(\lambda) = -R_\text{ROUGE}(\lambda)$, and $\delta = 0.1$. After MHT, all the schemes choose the hyperparameter $\lambda \in \hat{\Lambda}_\mathcal{Z}$ with the maximum estimated value $R_\text{ROUGE}(\lambda)$, i.e., minimum $\hat{R}_2(\lambda|\mathcal{Z})$ for LTT, and minimum $\hat{R}_2(\lambda|\mathcal{Z}_\text{OPT})$ for PT and RG-PT. The data set sizes are $|\mathcal{Z}| = 400$,  $|\mathcal{Z}_\text{OPT}| = 200$, and $|\mathcal{Z}_\text{MHT}| = 200$.

Fig. \ref{fig:seq2seq} illustrates the ROUGE-L score achieved on the test data by each calibration method, plotted against the target value for the BLEU score. The results demonstrate that RG-PT consistently maintains higher ROUGE-L scores, even under stricter requirements for the BLEU score. This highlights RG-PT's advantage in effectively exploring the hyperparameter space, enabling a more efficient testing procedure and identifying superior hyperparameter configurations that still statistically satisfy the desired conditions on the risk functions.

\section{Conclusion, Limitations, and Future Work}
\label{sec:conclusion}

In this paper, we have introduced RG-PT, a novel framework for multi-objective hyperparameter selection that integrates MHT with the concept of RGs to capture interdependencies among candidate hyperparameters. By leveraging a DAG structure informed by prior knowledge and data, RG-PT enables a more powerful parallel testing of hyperparameters compared to the state-of-the-art methods LTT and PT. RG-PT provides statistical guarantees through FDR control, while expanding the space of reliable hyperparameter configurations, leading to a superior optimization of auxiliary objectives.

Limitations of this work include the exclusive applicability to settings with discrete hyperparameter spaces and the lack of theoretical results on the power and sample efficiency of the method. Future work may focus on optimizing the RG structure to maximize power, on the use of synthetic data for the derivation of an RG, as well as on the integration with sequential testing methods based on e-processes \cite{zecchin2024adaptive}.

\section*{Acknowledgments}

This work was supported by the European Union’s Horizon Europe project CENTRIC (101096379), by the Open Fellowships of the EPSRC (EP/W024101/1), and by the EPSRC project (EP/X011852/1).


\newpage
\appendix

\section{Fixed Sequence Testing}
\label{appendix:FST}

In this section, we provide a brief overview of FST for controlling the FDR, which we used in our simulations for PT. While PT, as outlined in \cite{laufer2022efficiently}, is designed to support control of the FWER, our focus in this paper is on FDR control.

 MHT methods such as the Bonferroni correction for FWER control \cite{rice2007mathematical} and the Benjamini-Yekutieli (BY) procedure for FDR control \cite{benjamini2001control} do not leverage any side information about the relative reliability of the hyperparameters. When such information is available during calibration, FST can be used to test hyperparameters in order of expected reliability. When the ordering information is accurate, FST can be beneficial to reduce the FDR \cite{lynch2017control}. In this section, we briefly describe the FST procedure for FDR control.

 With FST, the candidate hyperparameters are ordered as \(\lambda_{(1)}, \ldots, \lambda_{(|\Lambda|)}\) using side information. The ordering ideally lists the hyperparameters from the most to the least likely to meet the reliability criterion (\ref{eq:lambda_reliable}).

 Starting with \( i = 1 \), each hyperparameter \( \lambda_{(i)} \) is tested sequentially based on its p-value \( p_{\lambda_{(i)}} \) against an adjusted critical value $\delta_i$ that decreases with the index $i = 1,2, \ldots, |\Lambda|$. At each step $i$, the hyperparameter $\lambda_{(i)}$ is deemed to be reliable if \( p_{\lambda_{(i)}} \leq \delta_i \). Testing continues until $k$ hyperparameters are deemed to be unreliable, at which point testing stops. The choice of the integer \( k \) is typically set as a small proportion, often around 5-10\%, of the total number of hypotheses, $|\Lambda|$.

 The critical values \( \delta_i \) are adapted to account for the position of each hypothesis in the testing sequence. These values are specifically designed to control the FDR under various dependency structures among the \( p \)-values. For the case of interest here, which is arbitrary dependence of the p-values, the critical levels can be set as \cite{lynch2017control}
\begin{equation}
\delta_i = \begin{cases}
\frac{\delta}{k} &\quad \text{if} \quad i \leq k \\
\frac{(|\Lambda| - k +1 )\delta}{(|\Lambda| - i +1 )k} &\quad \text{if} \quad i>k,
\end{cases}
\end{equation}
where \( \delta \) is the target FDR level and \( k \) is the number of unreliable hyperparameters allowed before testing stops.

 The final set of reliable hyperparameters is \( \hat{\Lambda}_\mathcal{Z} = \{\lambda_{(1)}, \ldots, \lambda_{(j)}\} \), where \( j \) corresponds to the index of the last hyperparameter tested before stopping. This ensures that the FDR is rigorously maintained below level $\delta$.

\section{Summary of DAGGER}
\label{appendix:DAGGER}

This section outlines the step-up procedure used in DAGGER \cite{ramdas2017dagger} to dynamically adjust the testing thresholds. DAGGER determines the testing thresholds adaptively based on the structure of the DAG and on the outcomes of previously tested hypotheses. At each depth level of the DAG, thresholds are updated dynamically to control the FDR, while respecting the hierarchical dependencies encoded by the DAG.

At each depth $d$, only the hyperparameters with no unreliable parents are considered for testing. If any parent of a hyperparameter $\lambda$ is deemed unreliable by DAGGER, all of its descendants are also automatically deemed unreliable. The threshold for testing the \( i \)-th hypothesis at depth \( d \) is given by
\begin{equation}
\label{eq:reshaped-step-up}
\delta_i(r) = \frac{v_i}{V} \cdot \frac{\delta}{\beta(m_i + r + R_{1:d-1} - 1)},
\end{equation}
where $r \in [1, |\Lambda_d|]$ is a parameter set as detailed below; \( v_i \) is the effective number of leaves in the subgraph rooted at the current node; \( V \) is the total number of leaves in the DAG; \( m_i \) is the effective number of nodes in the subgraph rooted at the current node; \( R_{1:d-1} \) is the total number of rejections at depths \( 1 \) through \( d-1 \); and \( \beta(\cdot) \) is a reshaping function, such as the Benjamini-Yekutieli \cite{benjamini2001control} function \( \beta_{BY}(x) = x /\sum_{k=1}^{V} 1/k \), which is designed to ensure FDR control under arbitrary dependence. 

The effective number of leaves $v_i$ and the effective number of nodes $m_i$ for node $i$ are calculated as follows. If $i$ is a leaf, then we have $v_i = m_i = 1$. Otherwise, the values $v_i$ and $m_i$ are calculated recursively from leaves to roots as 
\begin{equation}
    v_i = \sum_{j \in \text{children}(i)}\frac{v_j}{|\text{parents}(j)|},
\end{equation}
and 
\begin{equation}
    m_i = 1+\sum_{j \in \text{children}(i)}\frac{m_j}{|\text{parents}(j)|},
\end{equation}
where $\text{children}(i)$ and $\text{parents}(i)$ denote the sets of the children and the parents of node $i$, respectively.

The parameter $r$ at depth $d$ needs to be determined before testing can begin. To maximize the number of rejections while ensuring FDR control, DAGGER calculates the value
\begin{equation}
\label{eq:stepup}
    R = \arg \max_{r = 1, \ldots, |\Lambda_d|} \left[\sum_{\lambda_i\in \Lambda_d} \mathbf{1}\{p_{\lambda_i}(\mathcal{Z}_\text{MHT})\leq \delta_i(r) \} \geq r\right].
\end{equation}
The threshold $\delta_i(R)$ is then used to perform the testing for hyperparameter $\lambda_i$.

\begin{figure}
    \centering
    \includegraphics[width=\columnwidth]{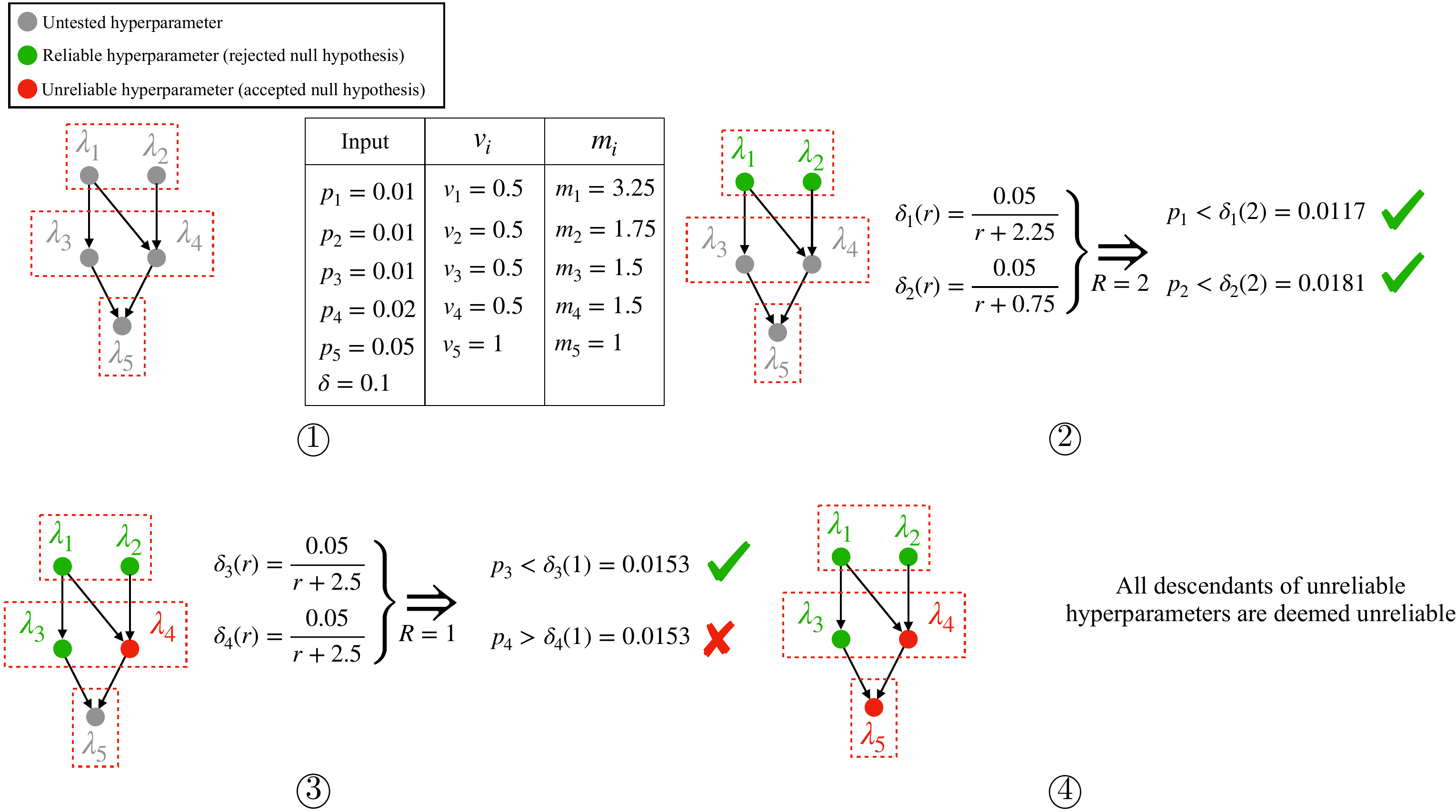}
   \caption{Illustration of the DAGGER algorithm's operation to control the FDR at $\delta = 0.1$. At each step, a hyperparameter is tested, starting from the root nodes and progressing level by level through the DAG. The testing thresholds $\delta_i$ are computed for each hyperparameter $\lambda_i$ using the step-up procedure in (\ref{eq:reshaped-step-up}) and (\ref{eq:stepup}), using the identity function as $\beta( \cdot )$. The p-value of each hyperparameter is compared against its respective threshold $\delta_i$ to assess the reliability of $\lambda_i$.}
    \label{fig:DAGGER}
\end{figure}

The step-up procedure (\ref{eq:reshaped-step-up}) ensures that thresholds are increasingly relaxed, i.e., increased, as we move further down the DAG. The overall algorithm is described in Algorithm \ref{alg:DAGGER}, and an example illustrating DAGGER’s operation is shown in Fig. \ref{fig:DAGGER}. The figure highlights how thresholds are updated and decisions propagated through the DAG. For simplicity, the figure assumes the reshaping function $\beta( \cdot )$ to be the identity function $\beta(x)=x$. First, the values for the effective number of leaves $v_i$ and the effective number of nodes $m_i$ are calculated for each hyperparameter $\lambda_i$. Next, going level by level, the thresholds $\delta_i(R)$ are calculated using (\ref{eq:reshaped-step-up}) and (\ref{eq:stepup}), and each hyperparameter $\lambda_i$ is tested by comparing $p_{\lambda_i}(\mathcal{Z}_\text{MHT})$ with $\delta_i(R)$.

\begin{algorithm}
    \caption{DAGGER \cite{ramdas2017dagger}}
    \label{alg:DAGGER}
    \begin{algorithmic}
        \STATE \textbf{Input:} DAG structure, p-values \( \{p_\lambda(\mathcal{Z}_\text{MHT})\} \), target FDR level \( \delta \)
        \STATE \textbf{Output:} Set of reliable hyperparameters \( \hat{\Lambda}_\mathcal{Z} \)

        \FOR{depth \( d = 1, \dots, D \)}
            \FOR{each hyperparameter \( \lambda_i\) in cluster \( \Lambda_d \)}
                \IF{all parent hyperparameters of $\lambda_i$ are deemed as reliable}
                \STATE Evaluate threshold $\delta_i(R)$ using (\ref{eq:reshaped-step-up}) and (\ref{eq:stepup})
                \IF{$p_{\lambda_i}(\mathcal{Z}_\text{MHT})\leq \delta_i(R)$}
                    \STATE Detect $\lambda_i$ as reliable
                \ELSE
                    \STATE Detect $\lambda_i$ as unreliable
                \ENDIF
                \ENDIF
            \ENDFOR
            \STATE Update \( \hat{\Lambda}_\mathcal{Z}\) with all the hyperparameters detected as reliable at depth $d$
        \ENDFOR
        \RETURN \( \hat{\Lambda}_\mathcal{Z} \)
    \end{algorithmic}
\end{algorithm}

\section{Computational Complexity of Constructing the Reliability Graph}

Constructing the RG involves three main steps: training the BT model, clustering, and Lasso regression. The training of the BT model has a time complexity $\mathcal{O}(n^2)$, because it involves a cyclic optimization over the model parameters associated to the pairs of configurations \cite[Sec.~3]{hunter2004mm}. Hierarchical clustering has a complexity $\mathcal{O}(n^2)$ \cite[Sec.~3]{mullner2011modern}, while Lasso regression has a per-iteration complexity $\mathcal{O}(n)$ \cite[Sec.~2.1]{friedman2010regularization}. Therefore, the overall complexity of RG construction is of the order $\mathcal{O}(n^2)$.

\section{RG-PT Algorithm}
\label{appendix:algorithm}

Algorithm \ref{alg:rgpt} provides a summary of RG-PT.

\begin{algorithm}[H]
\caption{Reliability Graph-Based Pareto Testing (RG-PT)}
\label{alg:rgpt}
\begin{algorithmic}[1]
\STATE \textbf{Input:} Hyperparameter set $\Lambda$, calibration dataset $\mathcal{Z}$, FDR level $\delta$, reliability thresholds $\{\alpha_l\}_{l=1}^{L_c}$, number of DAG levels $D$
\STATE \textbf{Output:}  Reliable hyperparameter subset $\hat{\Lambda}_\mathcal{Z}$

\STATE Split calibration data: $\mathcal{Z} = \mathcal{Z}_{\text{OPT}} \cup \mathcal{Z}_{\text{MHT}}$
\STATE Estimate Pareto front $\Lambda_{\text{OPT}} \subseteq \Lambda$ using $\mathcal{Z}_{\text{OPT}}$

\STATE \textbf{Construct Reliability Graph (RG):}
\STATE Compute pairwise comparisons using $\mathcal{Z}_{\text{OPT}}$ and optional priors
\STATE Estimate scores $s(\lambda)$ via Bradley-Terry model
\STATE Cluster $\Lambda_{\text{OPT}}$ into $D$ levels using scores
\FOR{each level $d = 2$ to $D$}
    \FOR{each $\lambda \in \Lambda_d$}
        \STATE Select parents in $\Lambda_{d-1}$ via non-negative Lasso
    \ENDFOR
\ENDFOR

\STATE \textbf{Run DAGGER on RG with data $\mathcal{Z}_{\text{MHT}}$:}
\FOR{each node $\lambda$ in topological order}
    \IF{all parents of $\lambda$ are reliable}
        \STATE Compute p-value $p_\lambda$
        \IF{$p_\lambda \leq \delta_\lambda$}
            \STATE Add $\lambda$ to $\hat{\Lambda}_\mathcal{Z}$
        \ENDIF
    \ENDIF
\ENDFOR

\RETURN $\hat{\Lambda}_\mathcal{Z}$
\end{algorithmic}
\end{algorithm}

\section{Additional Experiments}
\label{appendix:experiments}

In this section, we present four new experiments across language model calibration \cite{peters2021smoothing}, object detection \cite{angelopoulos2021learn}, image classification \cite{franceschi2024hyperparameteroptimizationmachinelearning}, and telecommunications engineering \cite{Nokia}, to demonstrate the advantages of our method over LTT and PT. We begin by detailing the hardware used for the simulations and the specific parameter settings chosen for each experiment.

\subsection{Experimental Setups}
\label{appendix:experimental_setup}

All experiments were conducted using dedicated computational resources. Specifically, RG-PT, LTT, and PT runs, along with data generation for the object detection, image classification, and telecommunications engineering tasks, were executed on a machine equipped with an Apple M1 Pro chip (10-core CPU, 16-core GPU, 16 GB RAM). Data generation for the prompt engineering experiment (Section~\ref{sec:prompt_engineering}) and the sequence-to-sequence translation task was performed on an NVIDIA A100 GPU (40 GB VRAM), using CUDA 11.3 and 40 GB system memory.

The RG-PT parameter settings for each experiment are summarized in Table~\ref{tab:rgpt-params}.

\begin{table}[h]
\centering
\caption{RG-PT parameter settings for each experiment.}
\label{tab:rgpt-params}
\begin{tabular}{lccc}
\toprule
\textbf{Experiment} & \textbf{$D$} & \textbf{$n_p$} & \textbf{$\tau$} \\
\midrule
Prompt Engineering  & 17 & 1000 & 0.1  \\
Sequence-to-sequence translation         & 10 & 200 & 0.1  \\
Object Detection                   & 20 & 0 & 0.1  \\
Image Classification - Low Dimension     & 10 & 0 & 0.1  \\
Image Classification - High Dimension & 20 & 0 & 0.1 \\
Telecommunications Engineering     & 10 & 0 & 0.1  \\
\bottomrule
\end{tabular}
\end{table}

\subsection{FDR Analysis Across All Experiments}

To begin with, we present a high-level summary of our experimental validation of Proposition~\ref{prop:prop}. The specific details of each experiment, including their objectives, datasets, and task configurations, are provided in their respective sections. Briefly, for each task, we ran RG-PT 100 times using different random splits of the available dataset into $\mathcal{Z}_\text{OPT}$, $\mathcal{Z}_\text{MHT}$, and an independent test set. The target was to control the average FDR on the test set below a threshold of $\delta = 0.1$. We then measured the average FDR on the test set across runs, and the results, summarized in Table~\ref{tab:fdr}, confirm that RG-PT satisfies the FDR condition as theoretically established in Proposition~\ref{prop:prop}.

\begin{table}[h]
\centering
\caption{Average FDR achieved by RG-PT across tasks for a target FDR threshold of $\delta = 0.1$.}
\label{tab:fdr}
\resizebox{\textwidth}{!}{%
\begin{tabular}{lccccc}
\toprule
\textbf{Task} & Prompt Engineering & Object Detection & Language Translation & Image Classification & Radio Access Scheduling \\
\midrule
\textbf{Average FDR} & 0.089 & 0.093 & 0.095 & 0.084 & 0.093 \\
\bottomrule
\end{tabular}%
}
\end{table}

\subsection{Ablation Study}
\label{appendix:ablation}

In this section, we use the prompt engineering task in Sec. \ref{sec:prompt_engineering} to perform an ablation study over the RG depth $D$, as well as the effect of misspecified prior information.

\subsubsection{Effect of DAG Depth}

\begin{figure}
    \centering
    \includegraphics[width=0.5\linewidth]{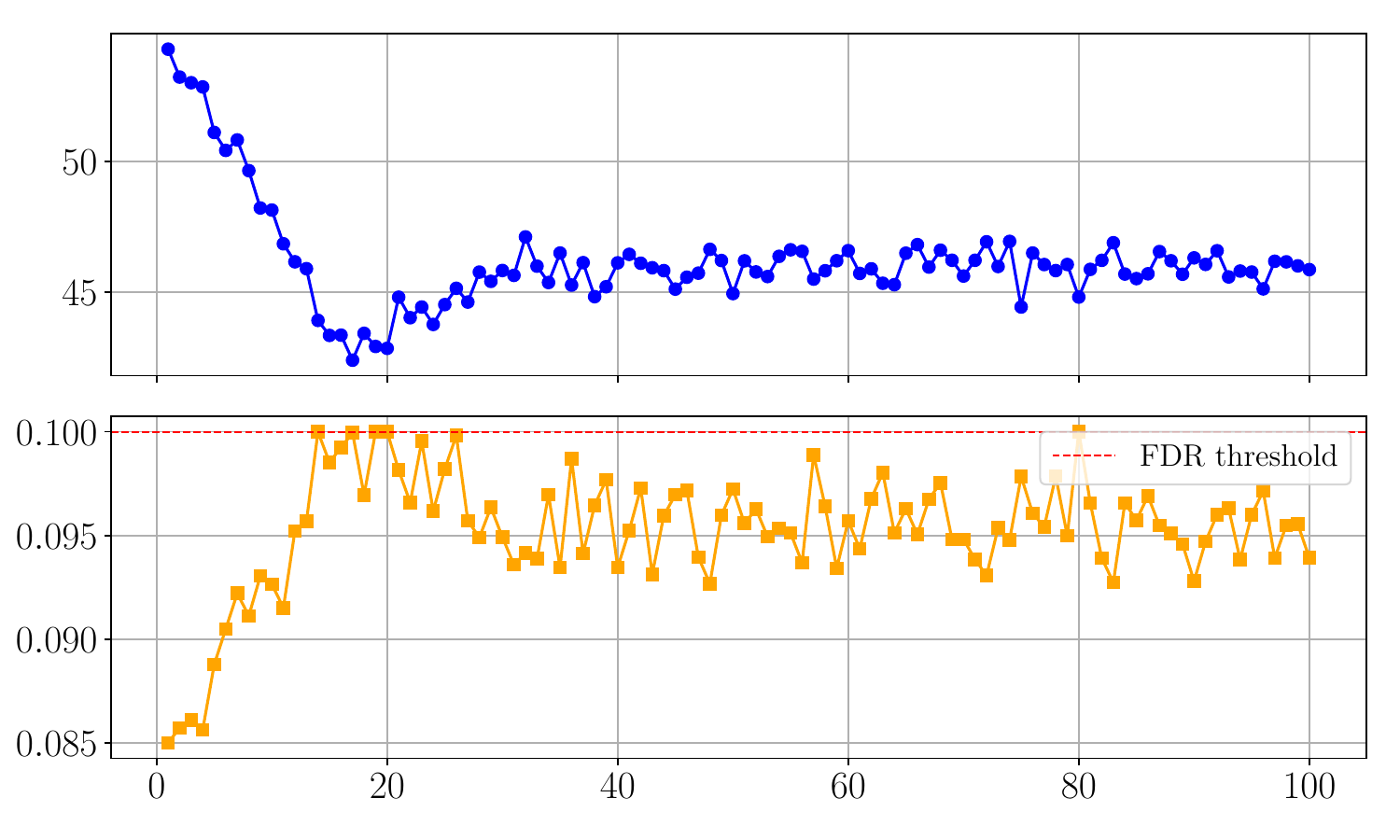}
    \caption{Average shortest prompt length in the returned prompt set $\hat{\Lambda}_\mathcal{Z}$ and the average FDR achieved by RG-PT as a function of RG depth $D$.}
    \label{fig:depth}
\end{figure}

To assess the impact of the DAG depth $D$ in RG-PT, we vary it from 1 (flat graph, equivalent to LTT) to 100 (fully linear, equivalent to PT), and measure both the length of the shortest prompt in $|\hat{\Lambda}_\mathcal{Z}|$ and the test FDR. Fig.~\ref{fig:depth} illustrates the average shortest reliable prompt length and average FDR over 100 runs for each depth, showing that FDR remains valid across depths. Additionally, it can be seen that there exists an intermediate depth that minimizes the average prompt length, indicating that the depth $1<D<|\Lambda_\text{OPT}|$ can be chosen to optimize power.

\subsubsection{Effect of Misspecified Priors}

\begin{figure}
    \centering
    \includegraphics[width=0.5\linewidth]{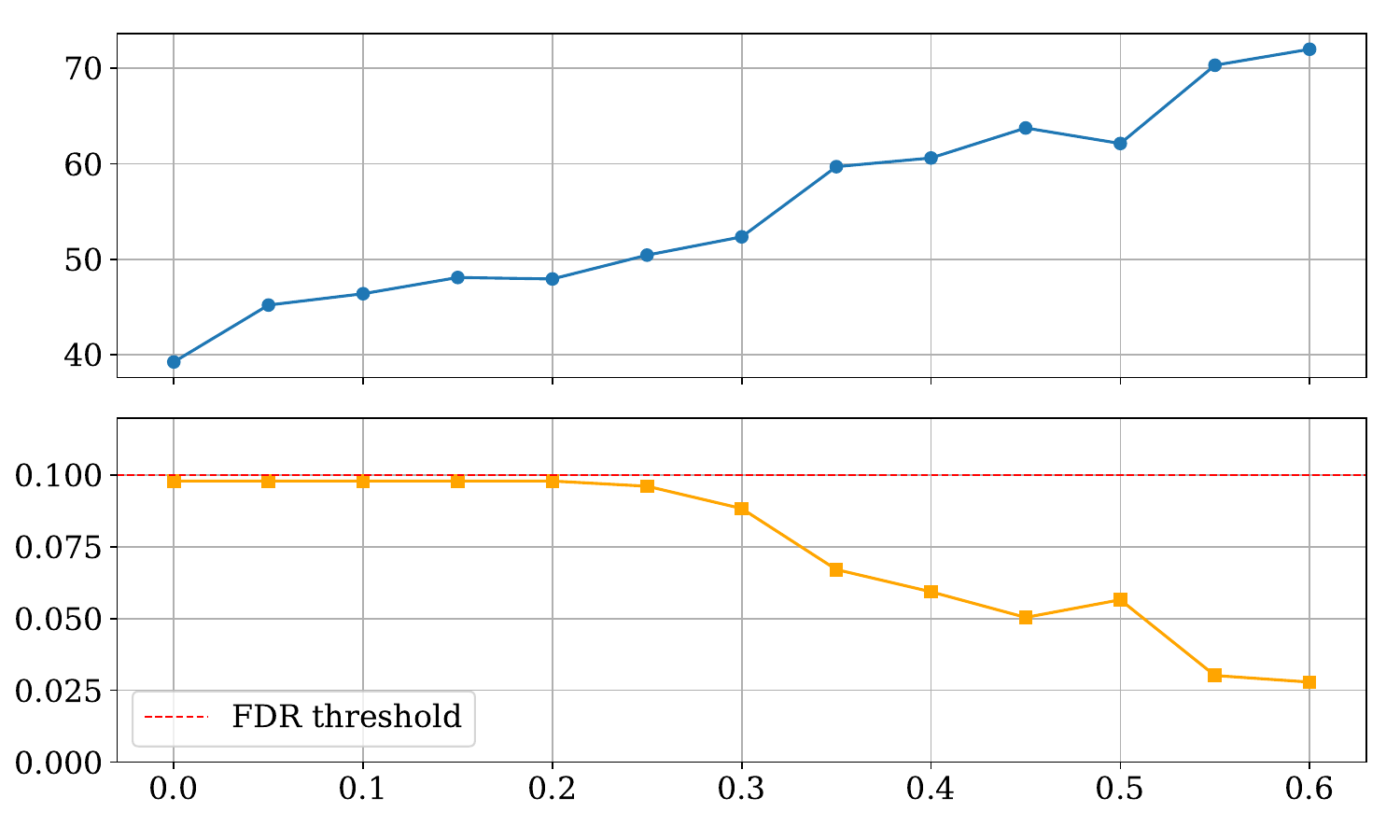}
    \caption{Average shortest prompt length in the returned prompt set $\hat{\Lambda}_\mathcal{Z}$ and the average FDR achieved by RG-PT as a function of fraction $f$ of flipped pairwise prior probabilities.}
    \label{fig:wrong}
\end{figure}

To simulate prior misspecification, we inject noise into the pairwise priors \(\eta_{ij}\). For every pair \(\lambda_i,\lambda_j\in\Lambda_\text{OPT}\), we independently swap the priors \(\eta_{ij}\) and \(\eta_{ji}\) with probability \(f\in[0,1]\). 
When \(f=0\) the prior remains intact, and when \(f=1\) every pairwise preference is completely reversed.

Fig.~\ref{fig:wrong} illustrates the average shortest prompt length in the returned set $\hat{\Lambda}_\mathcal{Z}$ of reliable prompts and the average FDR, across 100 random data splits, as a function of the flipped fraction $f$ of pairwise preferences $\eta_{ij}$. As $f$ increases, the shortest prompt length grows, indicating the effect of reduced prior information quality. Despite this, the FDR remains controlled below the target level $\delta = 0.1$, demonstrating robustness to misspecified priors. Notably, for $f > 0.6$, RG-PT returns an empty set $\hat{\Lambda}_\mathcal{Z}$ across all runs, correctly avoiding any potentially unreliable hyperparameters in the face of highly corrupted priors.

\subsection{Image Segmentation for Object Detection}
\label{sec:object_detection}

We now evaluate the proposed RG-PT framework on a multi-objective image segmentation task for object detection, leveraging the MS-COCO dataset \cite{lin2014microsoft} and a pretrained detector from Detectron2 \cite{detectron2} as done in \cite{angelopoulos2021learn}. The task involves three distinct objectives: (\textit{i}) detecting objects within an image (object detection); (\textit{ii}) delineating object boundaries (image segmentation); and (\textit{iii}) assigning correct labels to detected objects (object classification). These tasks are measured using recall, intersection-over-union (IoU), and classification accuracy, respectively. The goal is to control classification errors while optimizing recall and segmentation quality, addressing the trade-offs among these objectives.

The performance of the detection is determined by three hyperparameters:

    1. The \textit{object recall threshold} ($\lambda_1$) controls the threshold for selecting objects based on confidence scores. Reducing the value of $\lambda_1$ lowers the confidence threshold, which allows more objects to be selected at the cost of, potentially, increasing false positives.
    
    2. The \textit{mask size threshold} ($\lambda_2$) tunes the size of the bounding masks used to segment objects, impacting the IoU score.
        
    3. The \textit{classification certainty level} ($\lambda_3$) controls the certainty level required for object classification, adjusting the tolerance for inclusion in the set of labels assigned to each detected object.

The candidate hyperparameter set $\Lambda$ was constructed as per \cite{angelopoulos2021learn}, by taking all combinations of 50 linearly spaced values in $[0.2, 0.5]$ for $\lambda_1$, 5 linearly spaced values in $[0.3, 0.7]$ for $\lambda_2$, and 25 logarithmically spaced values in $[-0.00436, 0]$ for $\lambda_3$. These discretization choices were optimized \cite{angelopoulos2021learn}.

\begin{figure}
    \centering
    \includegraphics[width=0.5\linewidth]{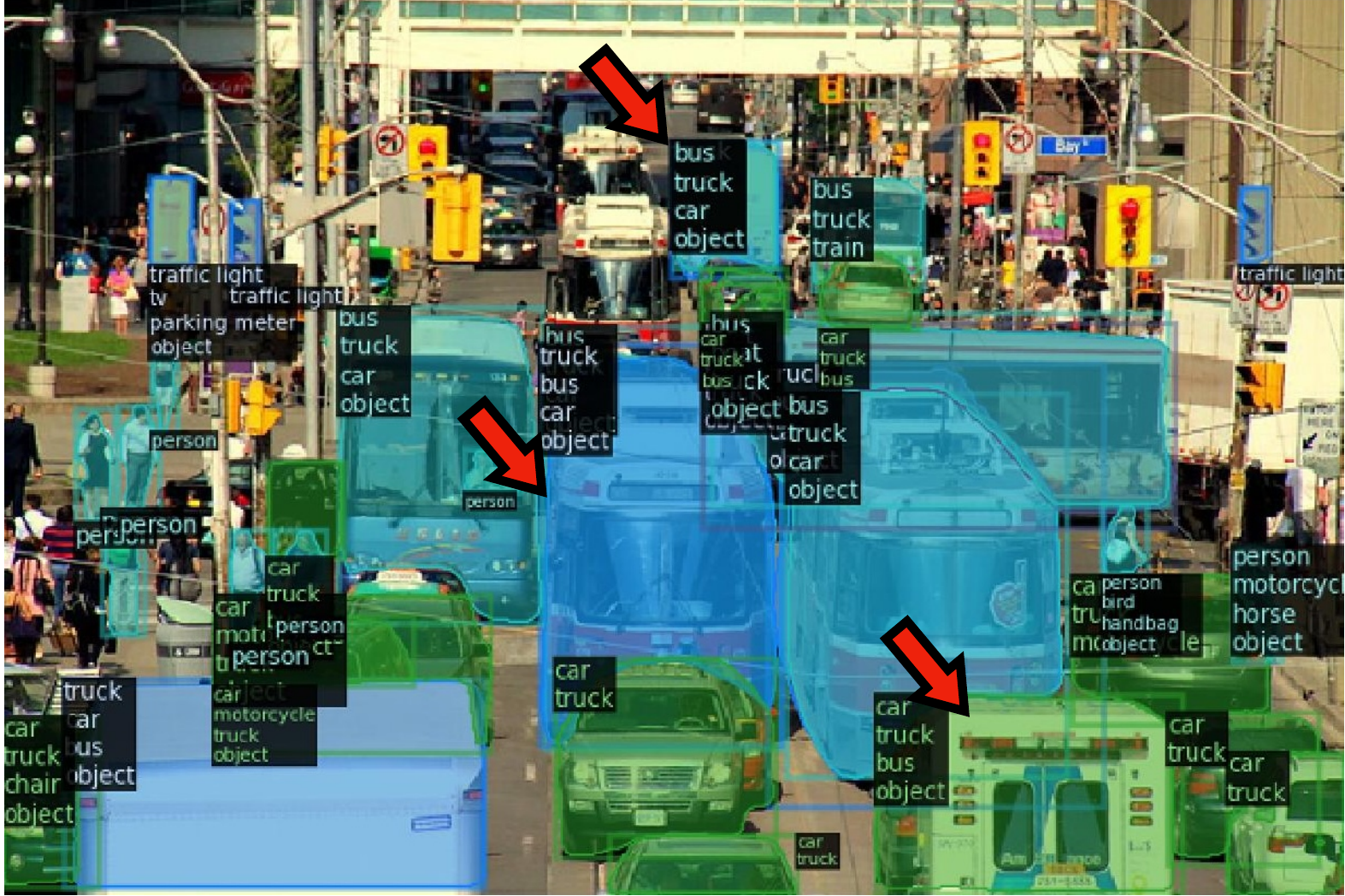}
    \caption{Illustration of the benefits of the proposed RG-PT hyperparameter selection scheme over the state-of-the-art LTT and PT for an object detection application \cite{angelopoulos2021learn}. The red arrows mark the objects not detected by an object recognition model calibrated using LTT or PT that are instead detected by the same model calibrated via RG-PT (see Appendix \ref{sec:object_detection} for details).}
    \label{fig:object_detection_comparison}
\end{figure}

\begin{figure}
    \centering
    \includegraphics[width=0.5\linewidth]{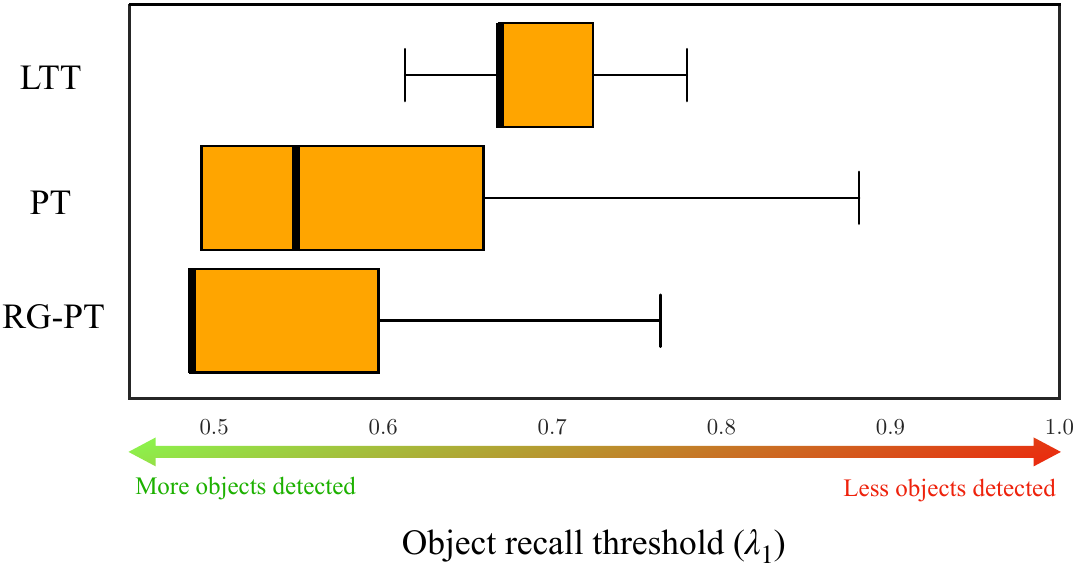}
    \caption{Example of of hyperparameter distributions for LTT, PT, and RG-PT methods. The box plots show the range (denoted by the horizontal black whiskers), median (represented by the thick black vertical lines), and interquartile range (depicted by the boxes) of the object recall threshold hyperparameter $\lambda_1$.}
    \label{fig:hyperparameter_comparison}
\end{figure}

 Denote as $R_1(\lambda)$, $R_2(\lambda)$, and $R_3(\lambda)$ the risks associated with recall, IOU, and coverage, respectively, for hyperparameter $\lambda = (\lambda_1, \lambda_2, \lambda_3)$. Controlling these risks in the context of problem (\ref{eq:goal}) is equivalent to having $L=3$ and $L_c = 3$. Additionally, as in \cite{angelopoulos2021learn}, we set the targets as $\alpha_1 = 0.5$, $\alpha_2 = 0.5$, and $\alpha_3 = 0.75$ with $\delta = 0.1$. Within the set $\hat{\Lambda}_\mathcal{Z}$ of reliable hyperparameters returned by the algorithm of choice, we choose the hyperparameter in subset $\hat{\Lambda}_\mathcal{Z}$ with the lowest value of $\lambda_1$ in order to increase the number of detected objects as much as possible. No prior knowledge was leveraged in creating the RG in this experiment by setting $n_p = 0$.

 We compare the distribution of the hyperparameters returned by LTT, PT, and RG-PT. The distribution is obtained by running 200 trials for each algorithm over different splits of calibration data $\mathcal{Z}$ into subsets $\mathcal{Z}_\text{OPT}$ and $\mathcal{Z}_\text{MHT}$ with $|\mathcal{Z}_\text{OPT}| = 1500$ and $|\mathcal{Z}_\text{MHT}| = 1500$. As shown in Fig. \ref{fig:hyperparameter_comparison}, the results demonstrate that RG-PT tends to return lower values for $\lambda_1$ than both LTT and PT. In particular, both the mean and dispersion for RG-PT are lower than those for LTT and PT. A lower threshold $\lambda_1$ allows the detector to select more objects, which directly enhances object recall, while still maintaining controlled levels of segmentation and classification accuracy (see also Fig. \ref{fig:object_detection_comparison}).

\subsection{Image Classification}
\label{sec:SVM}

 In this experiment, following \cite{franceschi2024hyperparameteroptimizationmachinelearning}, we consider the problem of hyperparameter selection for a support vector machine (SVM) model used to classify images from the Fashion MNIST dataset. The Fashion MNIST is a widely used benchmark for image classification, consisting of 70,000 grayscale images of 10 different clothing categories.

 We consider two risk functions ($L = 2$) in problem (\ref{eq:goal}), namely the classification error $R_\text{err}(\lambda)$ and the recall, $R_\text{rec}(\lambda)$. The classification error $R_\text{err}(\lambda)$ measures the proportion of incorrectly classified images out of the total number of samples. The recall measures the ability of a model to correctly identify all the relevant instances of each class. Accordingly, for each class, the recall is computed as the ratio of correctly identified instances of that class to the total number of actual instances of the same class in the dataset. The recall  $R_\text{rec}(\lambda)$ represents the average of the recall values across all classes. 

 With reference to problem (\ref{eq:goal}), we aim at minimizing recall, i.e. $R_2(\lambda) = R_\text{rec}(\lambda)$, while keeping the classification error rate below $0.3$, i.e. $R_1(\lambda) = R_\text{acc}(\lambda)$, $L_c = 1$, and $\alpha_1 = 0.3$. The goal is therefore defined as
\begin{equation}
\label{eq:goal_SVM}
    \min_{\lambda \in \Lambda}  R_\text{rec}(\lambda) \quad \text{subject to} \quad R_\text{acc}(\lambda) < 0.3 .
\end{equation}
This is a non-trivial problem since the accuracy maximizing model may not also optimize the recall \cite{powers2011evaluation}.

 The SVM model requires the selection of two hyperparameters \cite{franceschi2024hyperparameteroptimizationmachinelearning}. The regularization parameter, $C$, controls the desired trade-off between maximizing the margin and minimizing the classification error. Lower values of $C$ allow for a softer margin that can overlook some misclassification errors, while higher values enforce stricter classification error requirements. The kernel coefficient, $\gamma$, determines the impact of a single training example on the decision boundary, with higher values capturing finer details in the data set but risking overfitting. To create the initial candidate set $\Lambda$, we selected hyperparameters over a grid of 25 combinations, using five logarithmically spaced values in the intervals $[-3, 3]$ and $[-4, 1]$ for $C$ and $\gamma$, respectively. This selection is in line with approaches such as \cite{yogatama2015bayesian} for SVM hyperparameter selection.

 We used 5,000 data points for training the SVM, and used an additional 5,000 data points as calibration data $\mathcal{Z}$. The calibration data set $\mathcal{Z}$ was in turn divided into two groups of size 2,500, for the data sets $\mathcal{Z}_\text{OPT}$ and $\mathcal{Z}_\text{MHT}$, respectively.

\begin{figure}
    \centering
    \begin{subfigure}[t]{0.48\textwidth}
        \centering
        \includegraphics[width=\textwidth]{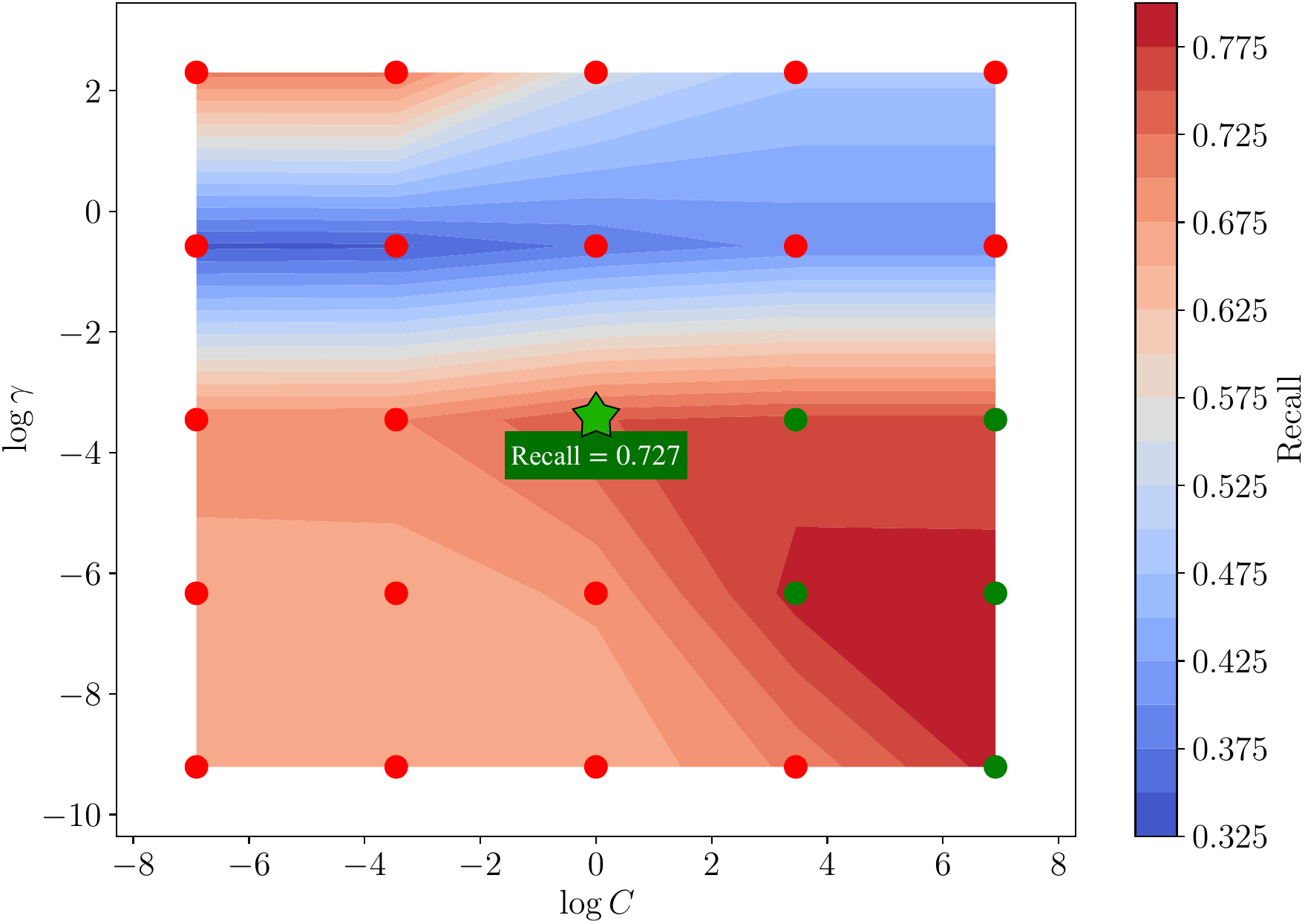}
        \caption{LTT}
        \label{fig:LTT_SVM}
    \end{subfigure}%
    \hfill
    \begin{subfigure}[t]{0.48\textwidth}
        \centering
        \includegraphics[width=\textwidth]{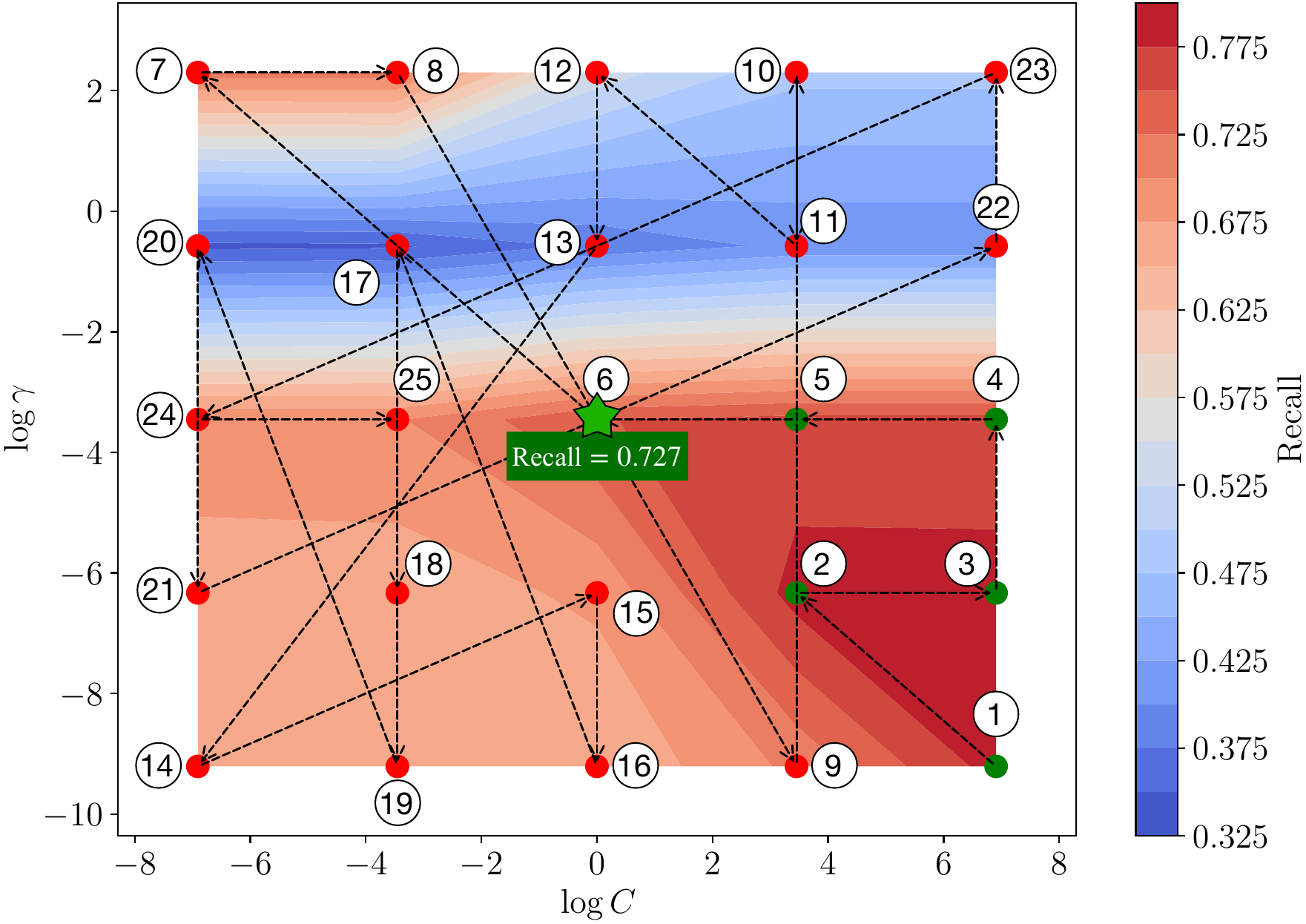}
        \caption{PT}
        \label{fig:Pareto_SVM}
    \end{subfigure}
    \hfill
    \begin{subfigure}[t]{0.48\textwidth}
        \centering
        \includegraphics[width=\textwidth]{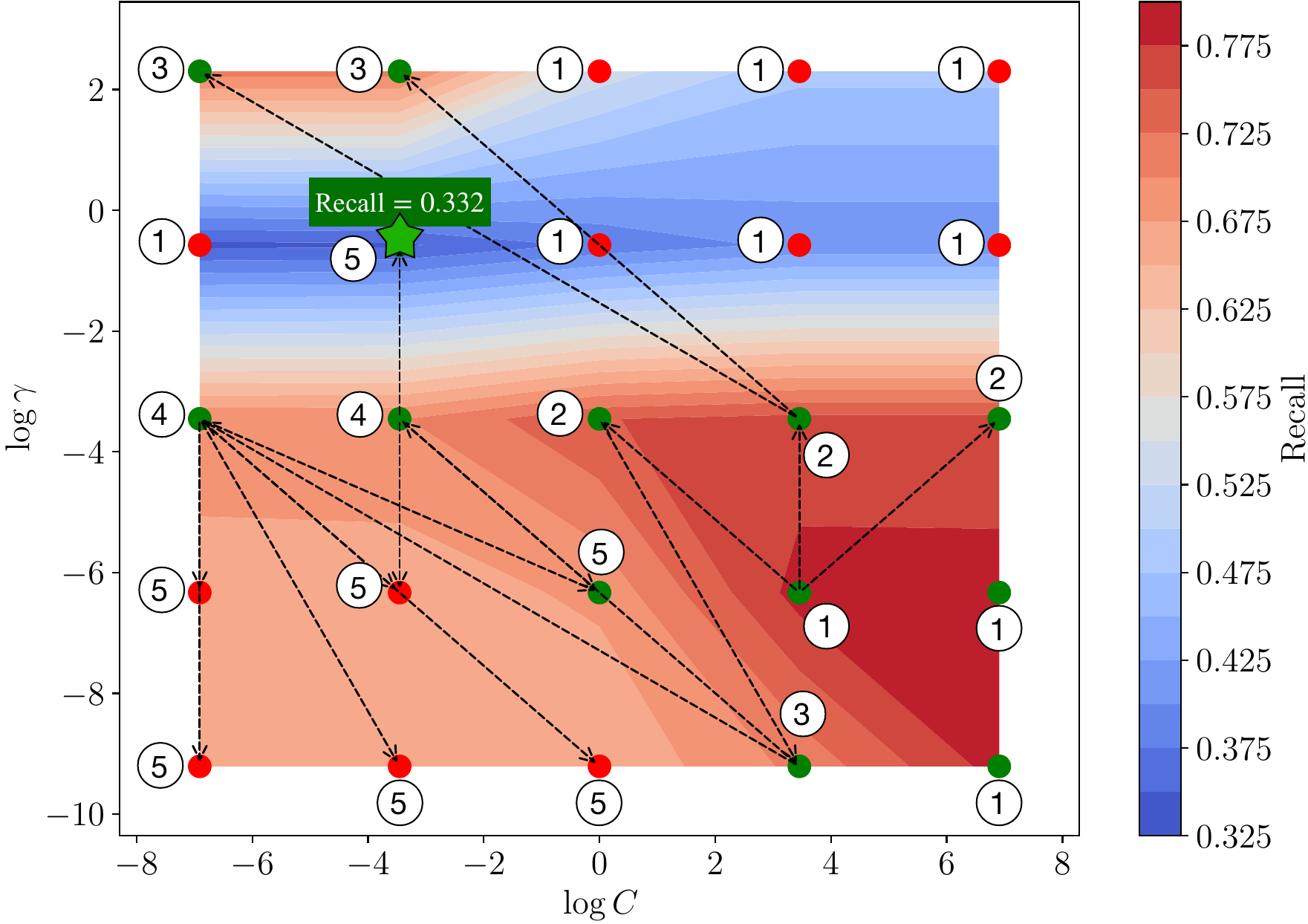}
        \caption{RG-PT}
        \label{fig:DAG_SVM}
    \end{subfigure}%
    \caption{Illustration of the hyperparameter selection procedure followed by LTT (a), PT (b), and RG-PT (c) for the setting studied in Sec. \ref{sec:SVM}. Each node represents a hyperparameter $\lambda = (C,\gamma)$, with the numbers representing the testing order. Green nodes show the hyperparameters included in the reliable set $\hat{\Lambda}_\mathcal{Z}$, and the star node shows the hyperparameter in set $\hat{\Lambda}_\mathcal{Z}$ with the lowest recall rate.}
    \label{fig:SVM}
\end{figure}

 Fig. \ref{fig:SVM} illustrates the testing procedures of LTT \cite{angelopoulos2021learn}, PT \cite{laufer2022efficiently}, and RG-PT. The x- and y-axes represent the logarithmic scales of the two hyperparameters $C$ and $\gamma$, while the contours indicate levels of recall $R_\text{rec}(\lambda)$, evaluated on the test data set, as a function of the hyperparameters $\lambda = (c, \gamma)$. The numbers illustrate the testing order for each testing method. Note that than LTT, which uses BY, does not follow any inferred order on the hyperparameters, and thus does not have the order labels in the figures. Furthermore, while LTT and PT test hyperparameters one by one, following a linear trajectory, RG-PT proceeds along a DAG, testing at the same time all hyperparameters at the same depth in the DAG. 

 LTT and PT are seen to stop at the sixth tested hyperparameter, yielding the set of reliable hyperparameters $\hat{\Lambda}_\mathcal{Z}$ marked as green dots. In contrast, RG-PT returns a much larger set $\hat{\Lambda}_\mathcal{Z}$ of reliable hyperparameters, also marked as green dots. Choosing within these sets the hyperparameter that minimizes the estimated recall as per problem (\ref{eq:goal_SVM}) yields the solutions indicated as green stars, corresponding to a test recall of 0.727 for LTT and PT, and 0.332 for RG-PT.

 It is important to note that all three methods yield test accuracies below the 0.3 threshold in (\ref{eq:goal_SVM}). Specifically, the hyperparameters selected by LTT and PT result in a test accuracy error of 0.267, while those chosen by RG-PT achieve a slightly higher accuracy error of 0.286. Although the accuracy of RG-PT is closer to the threshold, it remains consistent with the statistical guarantee outlined in (\ref{eq:goal_SVM}). In fact, RG-PT achieves a lower recall while maintaining the desired accuracy constraint, whereas LTT and PT follow a more conservative approach, leading to a reliable hyperparameter with higher recall.

To demonstrate the scalability of RG-PT to high-dimensional hyperparameter spaces, we repeated the previous experiment over a grid of 10,000 hyperparameter configurations instead of 25. This grid was constructed using 100 logarithmically spaced values for $C$ and $\gamma$ over the same intervals as before. The average FDR across 100 runs is reported in Table~\ref{tab:fdr}, highlighting RG-PT’s robustness and effectiveness even in high-dimensional settings.

\subsection{Radio Access Scheduling}

 In this section, we study a telecommunications engineering problem, namely the optimization of a radio access scheduler \cite{Nokia}. In this setup, each user equipment (UE) belongs to one of four quality-of-service (QoS) classes, assigned at random, each with its own delay and bit rate requirements \cite{de2020radio}. The goal is to control the delay of UEs in a given QoS class, while simultaneously minimizing the delays for UEs in the other three QoS classes. 

 Accordingly, in the context of problem (\ref{eq:goal}), we choose $L = 4$ and $L_c = 1$, and we set the risk $R_i(\lambda)$ to be equal to the average delay of QoS class $i$ for $1\leq i\leq 4$. We aim to keep $R_2(\lambda)$ below 15 ms, while minimizing $R_1(\lambda)$, $R_3(\lambda)$, and $R_4(\lambda)$. Formally, the problem is stated as 
\begin{equation}
\label{eq:goal_Nokia}
    \min_{\lambda \in \Lambda} \{R_1(\lambda), R_3(\lambda), R_4(\lambda)\} \; \text{subject to} \; R_2(\lambda) < 15 \; \text{ms} .
\end{equation}

 The scheduling algorithm at the base station allocates spectral resources to the UEs. As in \cite{Nokia}, the UEs are randomly distributed within a \(1 \, \text{km}^2\) area containing a centrally located base station. Each UE has an initial buffer of 100 packets, and moves at random speeds and directions. Resource allocation is carried out in intervals of 1 ms, called transmission time intervals (TTIs), over 10,000 TTIs per episode.

 The resource allocation algorithm is controlled by a set of hyperparameters \(\lambda = (\lambda_1, \lambda_2, \lambda_3, \lambda_4) \in \Lambda\), where each \(\lambda_i\) adjusts a specific criterion in the reward model as detailed in \cite{Nokia}. Hyperparameters $\lambda_1$, $\lambda_2$, $\lambda_3$, and $\lambda_4$ determine respectively the channel quality for each UE, the total queue sizes at the UEs, the age of the oldest packet in each UE’s buffer, and the fairness in resource block allocation among UEs.

 Calibration and test data were generated using the Nokia wireless suite \cite{Nokia}. For each run, we used 100 episodes for calibration and 100 episodes for testing. 

 The candidate hyperparameter set \(\Lambda\) was generated by keeping $\lambda_1$ and $\lambda_2$ at the values \(\lambda_1^*\) and  \(\lambda_2^*\) recommended by \cite{de2020radio}, and linearly sweeping hyperparameters $\lambda_3$ and $\lambda_4$ in [0.02, 0.2] and [-0.1, 0.1], respectively, with 10 steps each, resulting in a total of 100 combinations.

\begin{figure}
    \centering
    \includegraphics[width=0.9\columnwidth]{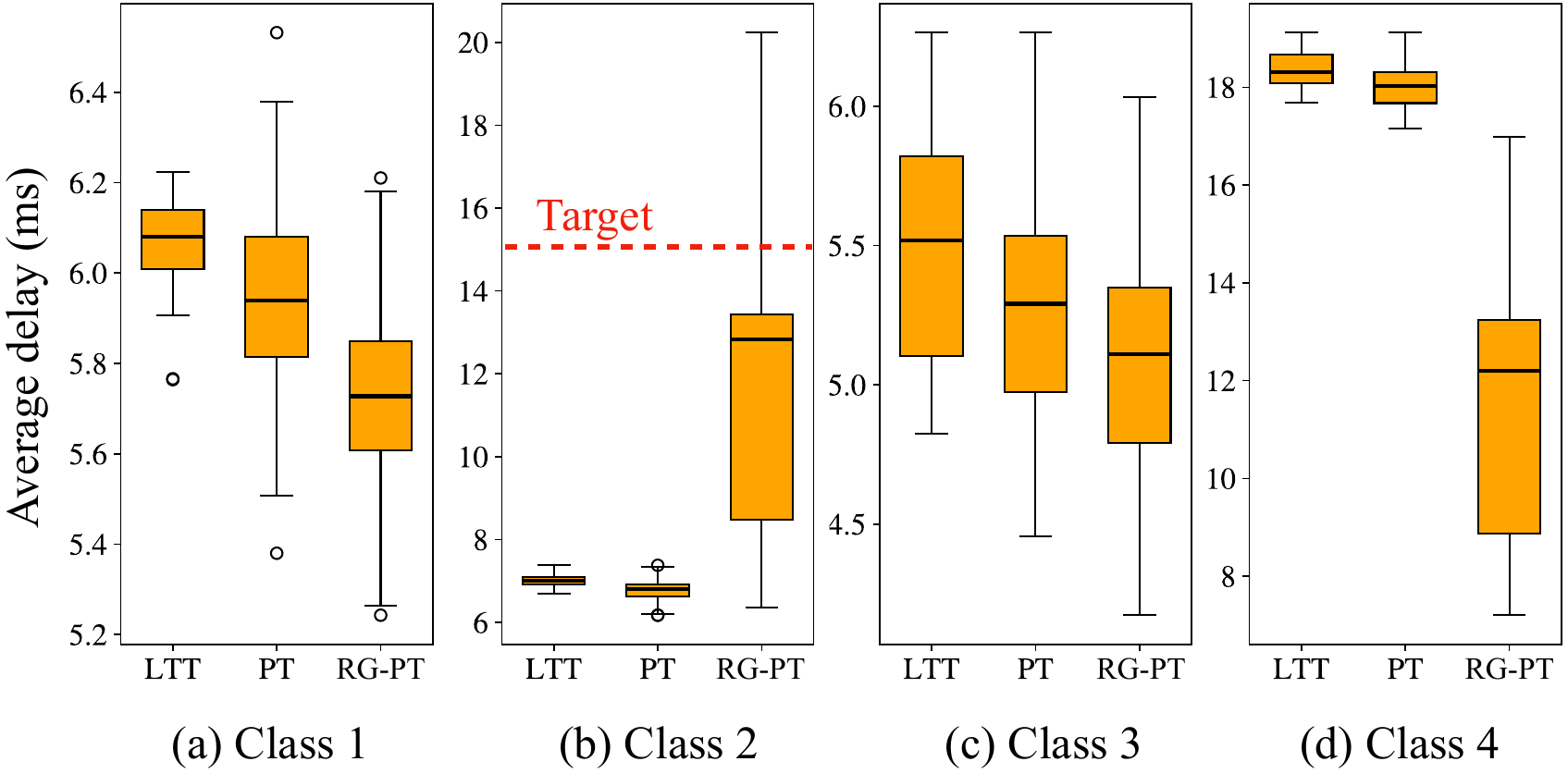}
    \caption{Distribution of the average delay for the four QoS classes using hyperparameters optimized by PT (left column) and RG-PT (right column). The dashed red line indicates the target threshold for the average delay in QoS class 2.}
    \label{fig:Nokia_dsitribution}
\end{figure}

 Fig. \ref{fig:Nokia_dsitribution} presents the results of using PT and RG-PT to optimize the hyperparameter $\lambda = (\lambda_1, \lambda_2, \lambda_3, \lambda_4)$. Both methods successfully meet the statistical guarantee of $R_2(\lambda) < 15$ ms for class 2. However, RG-PT demonstrates a greater ability to explore the hyperparameter space $\Lambda$, identifying configurations that more effectively minimize the average delay across the other three classes.

\end{document}